\documentclass{article}
\usepackage{microtype}
\usepackage{graphicx}
\usepackage{booktabs} 
\usepackage{hyperref}

\usepackage[accepted]{icml2018}

\usepackage{amsfonts}       
\usepackage{booktabs}       
\usepackage{cellspace}
\usepackage{amsmath}
\usepackage{framed}
\usepackage{graphicx}
\usepackage{makecell}
\usepackage{multirow}
\usepackage{subcaption}
\usepackage{tabularx}
\usepackage{tikz}
\usepackage{url}
\usepackage{wrapfig}
\usepackage{xspace}
\usepackage[toc,page]{appendix}
\usetikzlibrary{decorations.pathmorphing,shapes}

\xspaceremoveexception{-}
\newcommand{\name}{NetGAN\xspace}
\newcommand{\SBM}{DC-SBM\xspace}
\newcommand*{\ditto}{\multicolumn{1}{c}{*}}

\renewcommand{\cite}{\citep}
\def\bs{\ensuremath\boldsymbol}

\DeclareMathOperator*{\argmax}{arg\,max}
\DeclareMathOperator{\Cat}{Cat}

\DeclareMathOperator{\onehot}{onehot}

\DeclareMathOperator{\cov}{cov}

\begin{document}

\twocolumn[
\icmltitle{\name: Generating Graphs via Random Walks}

\icmlsetsymbol{equal}{*}

\begin{icmlauthorlist}
	\icmlauthor{Aleksandar Bojchevski}{equal,tum}
	\icmlauthor{Oleksandr Shchur}{equal,tum}
	\icmlauthor{Daniel Z\"ugner}{equal,tum}
	\icmlauthor{Stephan G\"unnemann}{tum}
\end{icmlauthorlist}

\icmlaffiliation{tum}{Technical University of Munich, Germany}

\icmlcorrespondingauthor{Daniel Z\"ugner}{zuegnerd@in.tum.de}

\icmlkeywords{Machine Learning, ICML}

\vskip 0.3in
]

\printAffiliationsAndNotice{\icmlEqualContribution} 

\begin{abstract}
We propose \name -- the first implicit generative model for graphs able to mimic real-world networks.
We pose the problem of graph generation as learning the distribution of biased random walks over the input graph.
The proposed model is based on a stochastic neural network that generates discrete output samples and is trained using the Wasserstein GAN objective.
\name is able to produce graphs that exhibit well-known network patterns without explicitly specifying them in the model definition. 
At the same time, our model exhibits strong generalization properties, as highlighted by its competitive link prediction performance, despite not being trained specifically for this task.
Being the first approach to combine both of these desirable properties, \name opens exciting avenues for further research.
\end{abstract}

\section{Introduction}
Generative models for graphs have a longstanding history, with applications including data augmentation, anomaly detection and recommendation \cite{chakrabarti2006graph}.
Explicit probabilistic models such as Barab\'asi-Albert or stochastic blockmodels are the de-facto standard in this field \cite{goldenberg2010survey}.
However, it has also been shown on multiple occasions that our intuitions about structure and behavior of graphs may be misleading.
For instance, heavy-tailed degree distributions in real graphs were in strong disagreement with the models existing at the time of their discovery \cite{barabasi1999emergence}.
More recent works like \citet{dong2017structural} and \citet{broido2018scale} keep bringing up other surprising characteristics of real-world networks that question the validity of the established models.
This leads us to the question: ``How do we define a model that captures all the essential (potentially still unknown) properties of real graphs?''

\begin{figure*}
	\centering
	\begin{subfigure}{0.33 \textwidth}
		\begin{tikzpicture}[decoration=snake]
		\node[anchor=south west,inner sep=0] (image) at (0,0) {\includegraphics[width= \textwidth]{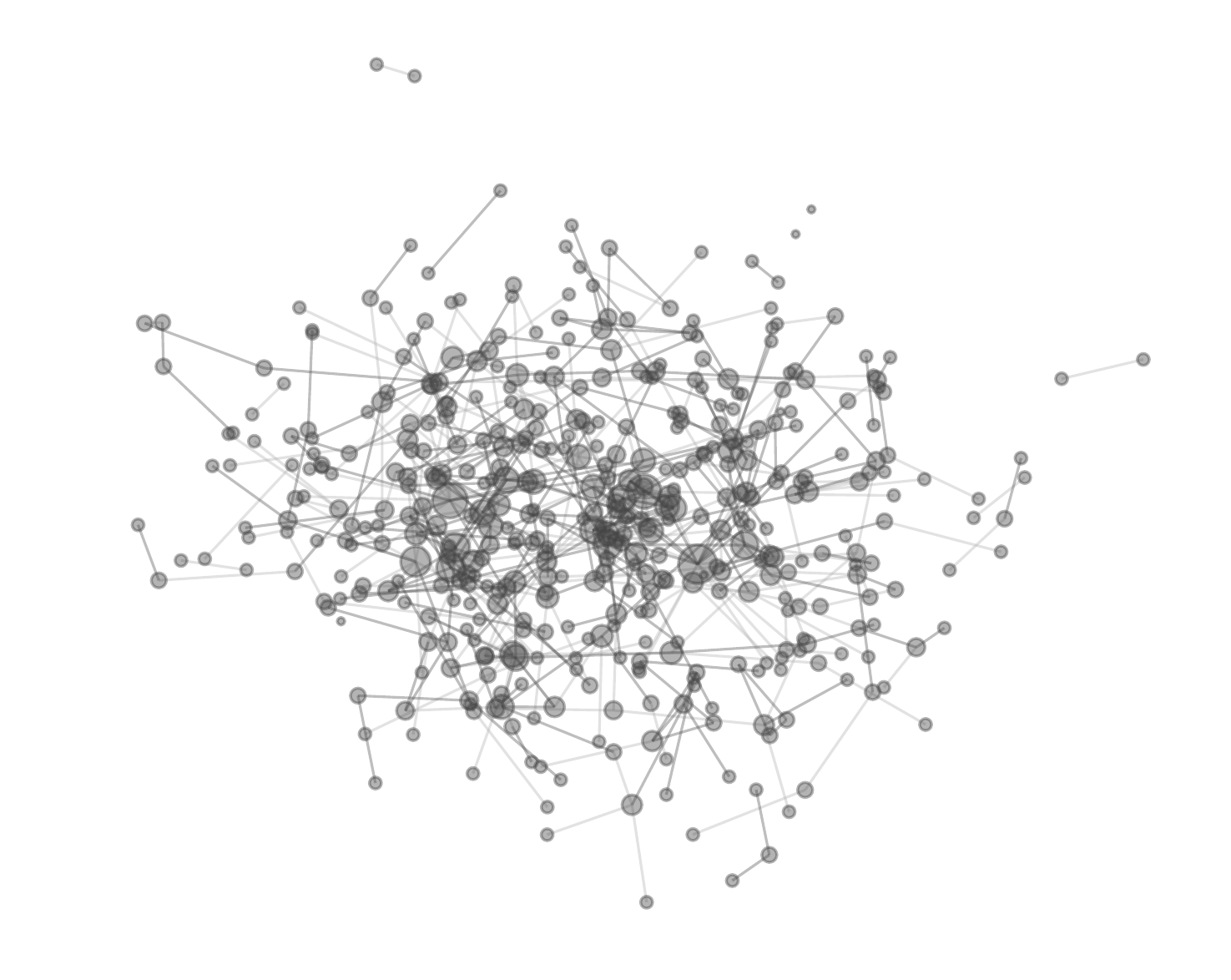}};
		\begin{scope}[x={(image.south east)},y={(image.north west)}] 
		\draw (0.64,0.72) circle (.4cm);
		\draw (0.31,0.26) circle (.4cm);
		\end{scope}
		\end{tikzpicture}
		\caption{Original graph}
		\label{sibling_a}
	\end{subfigure}
	\begin{subfigure}{0.33 \textwidth}
		\begin{tikzpicture}[decoration=snake]
		\node[anchor=south west,inner sep=0] (image) at (0,0) {\includegraphics[width= \textwidth]{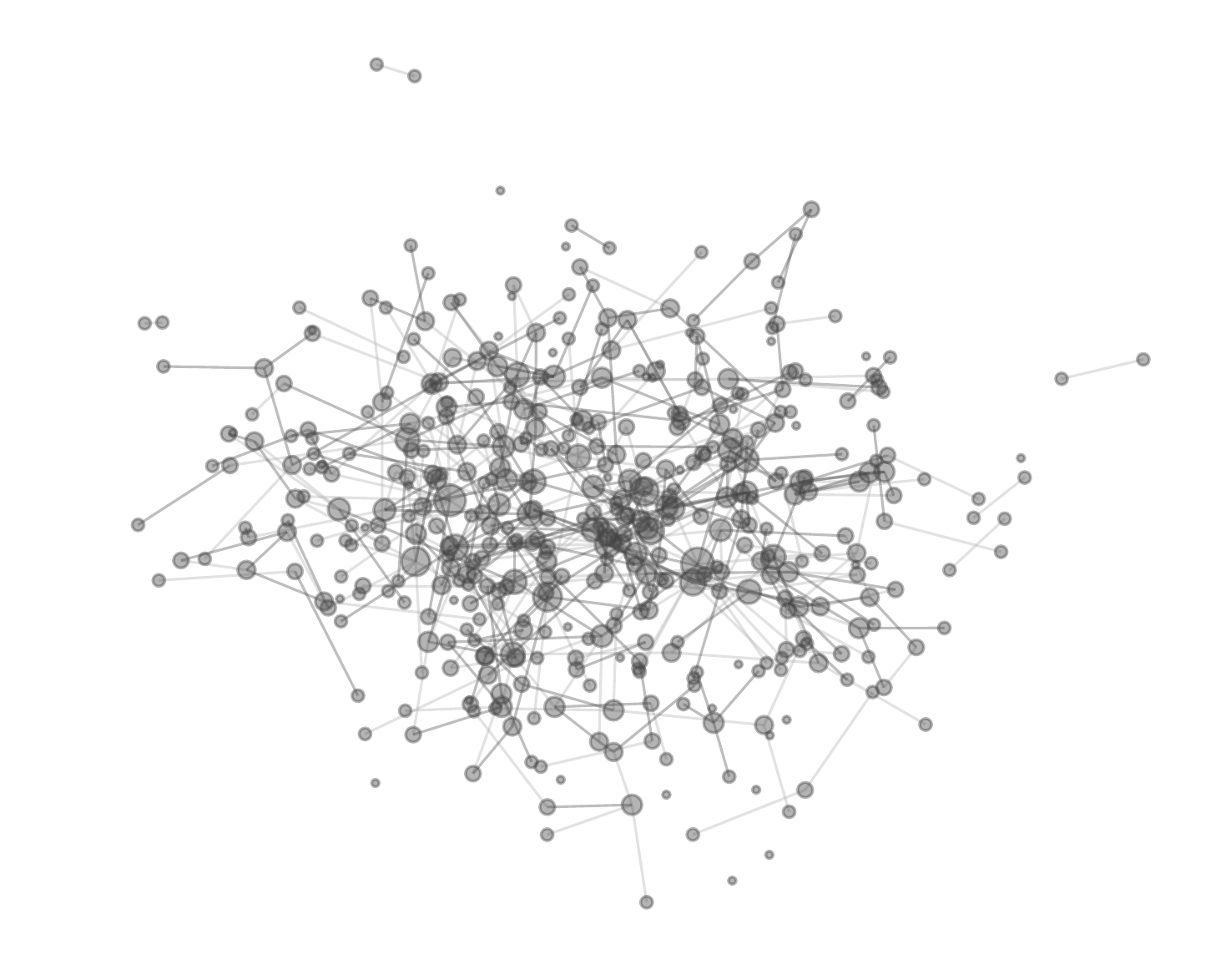}};
		\begin{scope}[x={(image.south east)},y={(image.north west)}] 
		\draw (0.64,0.72) circle (.4cm);
		\draw (0.31,0.26) circle (.4cm);
		\node[align=center, text width=1.1cm] at (.8, 0.1) {\baselineskip=5pt \scriptsize 44\%$\,$edge overlap \par} ;   
		
		\end{scope}
		\end{tikzpicture}
		\caption{Graph generated by \name}	
		\label{sibling_b}
	\end{subfigure} 
	\begin{subfigure}{0.324 \textwidth}
		\includegraphics[width=\textwidth]{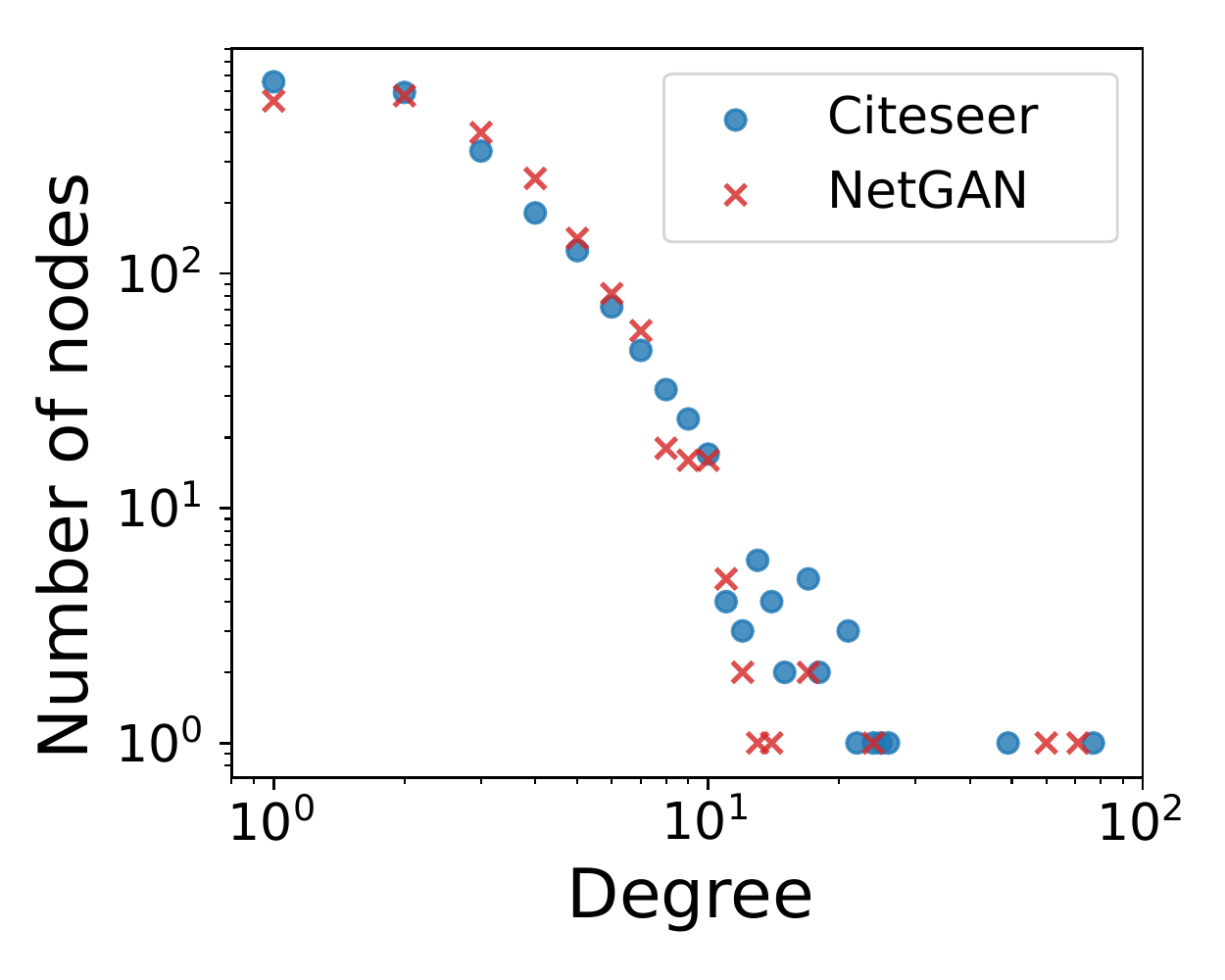}
		\caption{Degree distribution comparison}
		\label{sibling_c}
		
	\end{subfigure}
	
	\caption{(a) Subgraph of the \textsc{Citeseer} network and (b) the respective subset of the graph generated by \name. Both have similar structure but are not identical. (c) shows that the degree distributions of the two graphs are very close.}
	\label{fig:sibling_graph}
\end{figure*}
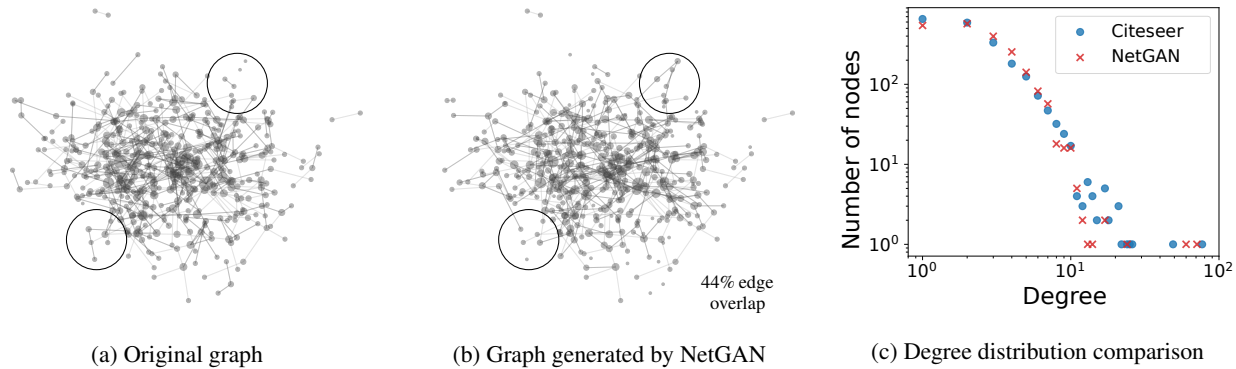

An increasingly popular way to address this issue in other fields is by switching from \emph{explicit} (prescribed) models to \emph{implicit} ones.
This transition is especially notable in computer vision, where
generative adversarial networks (GANs) \cite{goodfellow2014generative} significantly advanced the state of the art
{over the classic prescribed approaches like mixtures of Gaussians \cite{blanken2007multimedia}}.
GANs achieve unparalleled results in scenarios such as image and 3D objects generation \cite[e.g.,][]{karras2017progressive, berthelot2017began, wu2016learning}.
However, despite their massive success when dealing with real-valued data, adapting GANs to handle \emph{discrete} objects like graphs or text remains an open research problem \cite{goodfellow2016nips}.
In fact, discreteness is only one of the obstacles when applying GANs to network data.
Large repositories of graphs that all come from the same distribution are not available.
This means that in a typical setting one has to learn from a \emph{single graph}.
Additionally, any model operating on a graph necessarily has to be \emph{permutation invariant}, as graphs are isomorphic under node reordering.

In this work we introduce \emph{\name} -- the first implicit generative model for graphs and networks that tackles all of the above challenges.
We formulate the problem of learning the graph topology as learning the distribution of biased random walks over the graph.
Like in the typical GAN setting, the generator $G$ -- in our case defined as a stochastic neural network with discrete output samples -- learns to generate random walks that are \emph{plausible} in the real graph,
while the discriminator $D$ then has to distinguish them from the true ones that are sampled from the original graph.

The main requirement for a graph generative model is the ability to generate realistic graphs.
In the experimental section we compare \name to other established prescribed models on this task.
We observe that our proposed method consistently reproduces most known patterns inherent to real-world networks without explicitly specifying any of them in the model definition (e.g., degree distribution, as seen in Fig. \ref{fig:sibling_graph}).
However, a model that simply replicates the original graph would also trivially fulfill this requirement, which clearly isn't our goal.
In order to prove that this is not the case we examine the generalization properties of \name by evaluating its link prediction performance.
As our experiments show, our model exhibits competitive performance in this task and even achieves state-of-the-art results on some datasets.
This result is especially impressive, since \name is not trained explicitly for performing link prediction.
To summarize, our main contributions are:
\begin{itemize}
	\item We introduce \name
	\footnote{
		Code available at:
		\href{https://www.kdd.in.tum.de/netgan}
		{https://www.kdd.in.tum.de/netgan}
	} -- the first of its kind GAN architecture that generates graphs via random walks.
	Our model tackles the associated challenges of staying permutation invariant, learning from a single graph and generating discrete output.
	
	\item We show that our method preserves important topological properties, without having to explicitly specifying them in the model definition.
	Moreover, we demonstrate how latent space interpolation leads to producing graphs with smoothly changing characteristics.
	
	\item We highlight the generalization properties of \name by its link prediction performance
	that is competitive with the state of the art on real-word datasets, despite the model not being trained explicitly for this task.
\end{itemize}
\section{Related Work}
So far, no GAN architectures applicable to real-world networks have been proposed.
\citet{liu2017can} propose a GAN architecture for learning topological features of subgraphs.
\citet{tavakolilearning} apply GANs to graph data by trying to directly generate adjacency matrices.
Because their model produces the entire adjacency matrix -- including the zero entries -- it requires computations and memory quadratic in the number of nodes.
Such quadratic complexity is infeasible in practice, allowing to process only small graphs, with reported runtime of over 60 hours for a graph with only 154 nodes.
In contrast, \name operates on random walks -- it considers only the non-zero entries of the adjacency matrix efficiently exploiting the sparsity of real-world graphs -- and is readily applicable to graphs with thousands of nodes.

Deep learning methods for graph data have mostly been studied in the context of node embeddings \cite{perozzi2014deepwalk, grover2016node2vec, kipf2016variational}.
The main idea behind these approaches is that of modeling the probabilities of each individual edge's existence, $p(A_{uv})$, as some function of the respective node embeddings, $f(\bs{h}_u, \bs{h}_v)$, where $f$ is represented by a neural network.
The recently proposed GraphGAN \cite{wang2017graphgan} is another instance of such prescribed edge-level probabilistic models, where $f$ is optimized using the GAN objective instead of the traditional cross-entropy.
Deep embedding based methods achieve state-of-the-art scores in tasks like link prediction and node classification.
Nevertheless, as we show in Sec.\@ \ref{sec:generation}, using such approaches for generating entire graphs produces samples that don't preserve any of the patterns inherent to real-world networks.

Prescribed generative models for graphs have a long history and are well-studied. 
For a survey we refer the reader to \citet{chakrabarti2006graph} and \citet{goldenberg2010survey}. 
Typically, prescribed generative approaches are designed to capture and reproduce some predefined subset of graph properties (e.g., degree distribution, community structure, clustering coefficient).
Notable examples include the configuration model \cite{bender1978asymptotic, molloy1995critical},
variants of the degree-corrected stochastic blockmodel \cite{karrer2011stochastic, DBLP:conf/aaai/BojchevskiG18},
Exponential Random Graph Models \citep{holland1981exponential}, Multiplicative Attribute
Graph model \cite{kim2011modeling}, and the block two-level Erd{\H{o}}s-R{\'e}niy random graph model \cite{seshadhri2012bter}.
In Sec.~\ref{sec:experiments} we compare with some of these prescribed models on the tasks of graph generation and link prediction.

Due to the challenging nature of the problem, only few approaches able to generate discrete data using GANs exist.
Most approaches focus on generating discrete sequences such as text, with some of them using reinforcement learning techniques to enable backpropagation through sampling discrete random variables \cite{yu2017seqgan, kusner2016gans, li2017adversarial, liang2017recurrent}.
Other approaches modify the GAN objective to tackle the same challenge \citep{che2017maximum, hjelm2017boundary}. 
Focusing on non-sequential discrete data, \citet{choi2017generating} generate high-dimensional discrete features (e.g. binary indicators, counts) in patient records.
None of these methods have considered graph structured data.

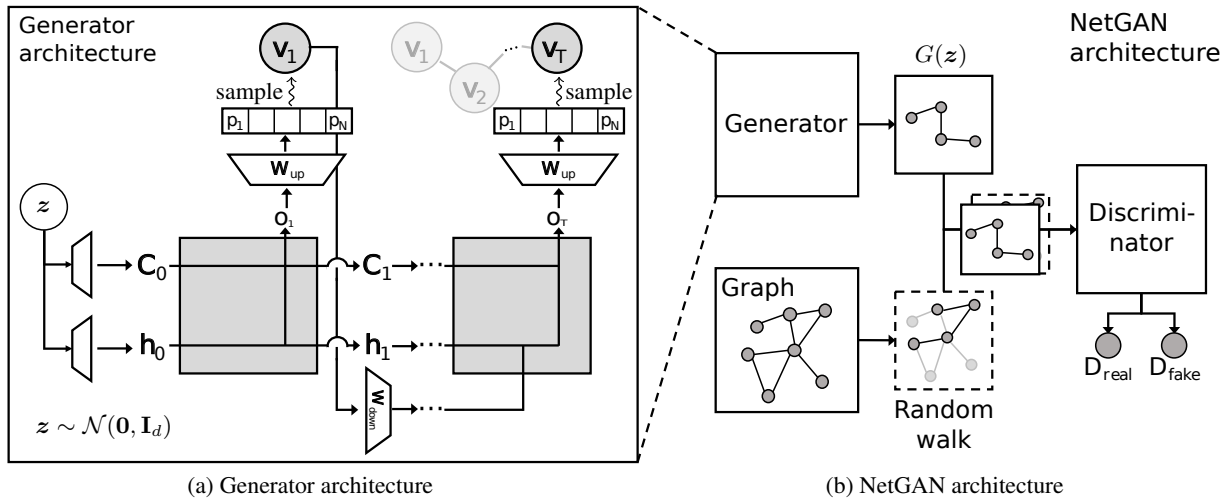
\begin{figure*}[ht!]
	\centering
	\begin{tikzpicture}[decoration=snake]
cc	\node[anchor=south west,inner sep=0] (image) at (0,0) {\includegraphics[width=0.94  \textwidth]{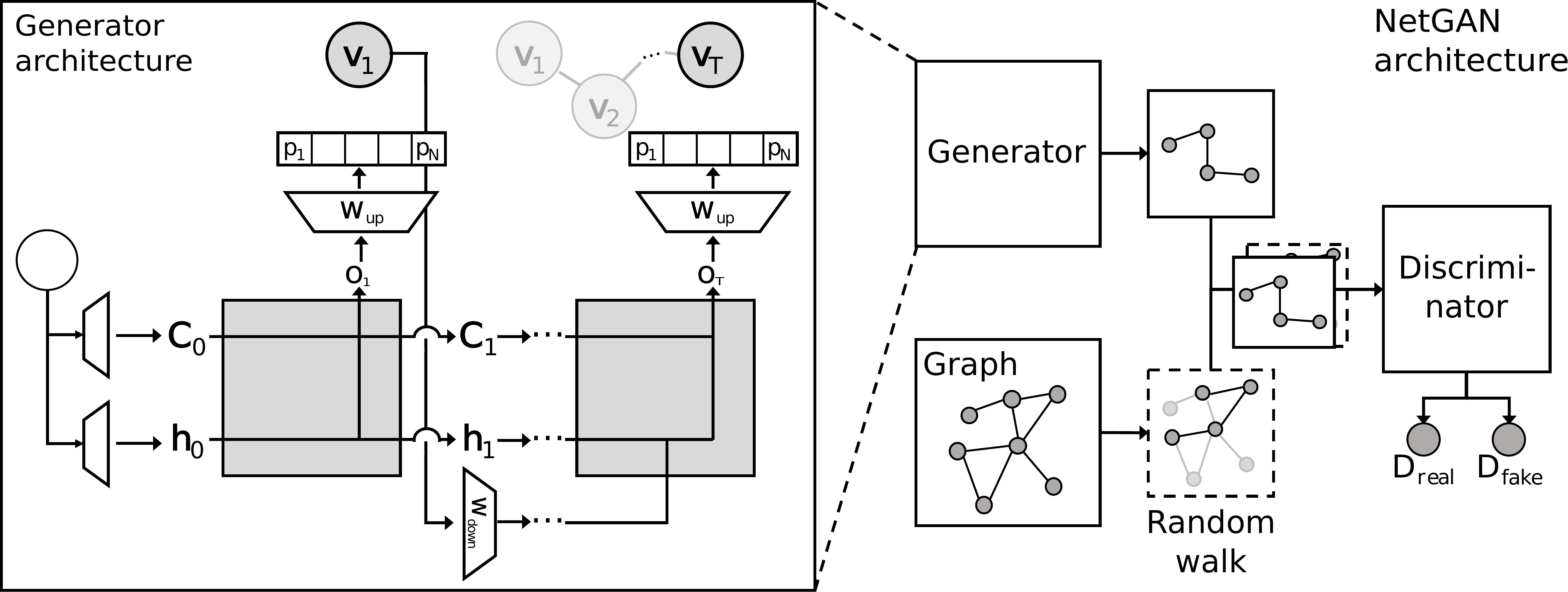}};
	\begin{scope}[x={(image.south east)},y={(image.north west)}] 
	\node[align=center] at (0.08, 0.08) {\small $\bs z \sim \mathcal{N}(\bs 0, \mathbf I_d)$};   
	\node[align=center] at (0.03, 0.56) {\small $\bs z$};   
	\node[align=center] at (0.77, 0.89) {\small $G(\bs z)$};   
	\node[align=right] at (0.20,0.807) {\small sample};   
	\draw [->,
	line join=round,
	line width=0.5 pt,
	decorate, decoration={
		snake,
		segment length=4,
		amplitude=.9,post=lineto,
		post length=2pt
	}]  (0.2345,0.78) -- (0.2345,0.844);
	
	\node[align=right] at (0.488,0.807) {\small sample};   
	\draw [->,
	line join=round,
	line width=0.5 pt,
	decorate, decoration={
		snake,
		segment length=4,
		amplitude=.9,post=lineto,
		post length=2pt
	}]  (0.455,0.78) -- (0.455,0.844);

	\node[align=right] at (0.25,-0.05) {\small (a) Generator architecture};
	\node[align=right] at (0.773,-0.05) {\small (b) \name architecture};      
	\end{scope}
	\end{tikzpicture}
	\vspace{-3mm}
	\caption{The \name architecture proposed in this work (b) and the generator architecture (a).}
	\label{fig:architecture_generator}
\end{figure*}

\section{Model}

In this section we introduce \name - a Generative Adversarial Network model for graph / network data.
Its core idea lies in capturing the topology of a graph by learning a distribution over the random walks.
Given an input graph of $N$ nodes, defined by a binary adjacency matrix $\bs A \in \{0, 1\}^{N \times N}$, we first sample a set of random walks of length $T$ from $\bs A$.
This collection of random walks serves as a training set for our model.
We use the biased second-order random walk sampling strategy described in \citet{grover2016node2vec}, as it better captures both local and global graph structure.
An important advantage of using random walks is their invariance under node reordering.
Additionally, random walks only include the nonzero entries of $\bs A$, thus efficiently exploiting the sparsity of real-world graphs.

Like any typical GAN architecture, \name consists of two main components - a generator $G$ and a discriminator $D$.
The goal of the generator is to generate synthetic random walks that are plausible in the input graph.
At the same time, the discriminator learns to distinguish the synthetic random walks from the real ones that come from the training set.
Both $G$ and $D$ are trained end-to-end using backpropagation.
At any point of the training process it is possible to use $G$ to generate a set of random walks, which can then be used to produce an adjacency matrix of a new generated graph.
In the rest of this section we describe each stage of this process and our design choices in more detail. An overview of our model's complete architecture can be seen in Fig. \ref{fig:architecture_generator}.

\subsection{Architecture}

\label{sec:architecture}
\textbf{Generator. }
The generator $G$ defines an implicit probabilistic model for generating random walks: $(\bs v_1, ..., \bs v_T) \sim G$.
We model $G$ as a sequential process based on a neural network $f_\theta$ parametrized by $\theta$.
At each step $t$, $f_\theta$ produces two values: the probability distribution over the next node to be sampled, parametrized by the logits
 $\bs p_t$, and the current memory state of the model, denoted as $\bs m_t$.
The next node $\bs v_t$, represented as a one-hot vector, is sampled from a categorical distribution $\bs v_t \sim \Cat(\sigma(\bs p_t))$, where $\sigma(\cdot)$ denotes the softmax function, and together with $\bs m_t$ is passed into $f_\theta$ at the next step $t+1$.
Similarly to the classic GAN setting, a latent code $\bs z$ drawn from a multivariate standard normal distribution is passed through a parametric function $g_{\theta^\prime}$ to initialize $\bs m_0$.
The generative process of $G$ is summarized in the box below. 
\begin{framed}
	\vspace{-7mm}
	\begin{align*}
	\label{eq:generator}
	&&\bs{z} \sim \mathcal{N}(\bs 0, \bs I_{d})\\
	&&\bs m_0 = g_{\theta^\prime}(\bs z)\\
	&\bs v_1 \sim \Cat(\sigma(\bs p_1)),  &(\bs p_1, \bs m_1) = f_{\theta}(\bs m_0, \bs 0)\\
	&\bs v_2 \sim \Cat(\sigma(\bs p_2)),  &(\bs p_2, \bs m_2) = f_{\theta}(\bs m_1, \bs v_1)\\
	&\quad\quad\vdots &\vdots\quad\quad\quad\quad\\
	&\bs v_T \sim \Cat(\sigma(\bs p_T)),  &(\bs p_T, \bs m_T) = f_{\theta}(\bs m_{T-1}, \bs v_{T-1})
	\end{align*}
	\vspace{-7mm}
\end{framed}
\vspace{-3mm}
In this work we focus our attention on the Long short-term memory (LSTM) architecture for $f_\theta$, introduced by \citet{lstm}.
The memory state $\bs m_t$ of an LSTM is represented by the cell state $\bs C_t$, and the hidden state $\bs h_t$.
The latent code $\bs z$ goes through two separate streams, each consisting of two fully connected layers with $\tanh$ activation, and then used to initialize $(\bs C_0, \bs h_0)$.

A natural question might arise: "Why use a model with memory and temporal dependencies, when the random walks are Markov processes?" {(2nd order Markov for biased RWs)}.
Or put differently, what's the benefit of using random walks of length greater than 2?
In theory, a model with large enough capacity could simply memorize all existing edges in the graph and recreate them. 
However, for large graphs achieving this in practice is not feasible. More importantly, pure memorization is not the goal of \name, rather we want to have generalization and to generate graphs with similar properties, not exact replicas. 
Having longer random walks combined with memory helps the model to learn the topology and general patterns in the data (e.g., community structure). 
Our experiments in Sec. \ref{sec:link_prediction} confirm this, showing that longer random walks are indeed beneficial.

After each time step, to generate the next node in the random walk, the network $f_\theta$ should output the logits $\bs p_t$ of length $N$. However,
operating in such high dimensional space leads to an unnecessary computational overhead.
To tackle this issue, the LSTM outputs $\bs o_t \in \mathbb{R}^{H}$ instead, with $H \ll N$, which is then up-projected to $\mathbb{R}^N$ using the matrix $\bs{W}_{up} \in \mathbb{R}^{H \times N}$. 
This enables us to efficiently handle large-scale graphs.

Sampling the next node in the random walk $\bs v_t$ presents another challenge. Since sampling from a categorical distribution is a non-differentiable operation it blocks the flow of gradients and precludes backpropagation. 
We solve this problem by using the Straight-Through Gumbel estimator by \citet{jang2016categorical}.
More specifically, we perform the following transformation: 
First, we let $\bs v^*_t = \sigma \left( (\bs p_t + \bs g)/\tau) \right)$, where $\tau$ is a temperature parameter, and $g_i$'s are i.i.d.\ samples from a Gumbel distribution with zero mean and unit scale.
Then, the next sample is chosen as $\bs v_t = \onehot(\argmax \bs v^*_t)$.
While the one-hot sample $\bs v_t$ is passed as input to the next time step,
during the backward pass the gradients will flow through the differentiable $\bs v^*_t$.
The choice of $\tau$ allows to trade-off between better flow of gradients (large $\tau$, more uniform $\bs v^*_t$) and more exact calculations (small $\tau$, $\bs v_t^* \approx \bs v_t$).

Now that a new node $\bs v_t$ is sampled, it needs to be projected back to a lower-dimensional representation before feeding into the LSTM.
This is done by means of down-projection matrix $\bs W_{down} \in \mathbb{R}^{N \times H}$.

\textbf{Discriminator. }
The discriminator $D$ is based on the standard LSTM architecture.
At every time step $t$, a one-hot vector $\bs v_t$, denoting the node at the current position, is fed as input.
After processing the entire sequence of $T$ nodes, the discriminator outputs a single score that represents the probability of the random walk being real.

\subsection{Training}\label{sec:training}

\textbf{Wasserstein GAN. }
We train our model based on the Wasserstein GAN (WGAN) framework \cite{arjovsky2017wasserstein}, as it prevents mode collapse and leads to more stable training overall.
To enforce the Lipschitz constraint of the discriminator, we use the gradient penalty as  in \citet{gulrajani2017improved}.
The model parameters $\{\theta, \theta^\prime\}$ are trained using stochastic gradient descent with Adam \cite{kingma2014adam}.
Weights are regularized with an $L_2$ penalty.

\textbf{Early stopping. }
Because we are interested in generalizing the input graph, the ``trivial'' solution where the generator has memorized all existing edges is of no interest to us.
This means that we need to control how closely the generated graphs resemble the original one.
To achieve this, we propose two possible early stopping strategies, either of which can be used depending on the task at hand.
The first strategy, named \textsc{Val-Criterion} is concerned with the generalization properties of \name.
During training, we keep a sliding window of the random walks generated in the last 1,000 iterations and use them to construct a matrix of transition counts. This matrix is then used to evaluate the link prediction performance on a validation set (i.e. ROC and AP scores, for more details see Sec. \ref{sec:link_prediction}). We stop with training when the validation performance stops improving.

The second strategy, named \textsc{EO-Criterion} makes \name very flexible and gives the user control over the graph generation. We stop training when we achieve a user specified edge overlap between the generated graphs (see next section) and the original one at a given iteration. Based on her end task the user can choose to generate graphs with either small or large edge overlap with the original, while maintaining structural similarity. This will lead to generated graphs that either generalize better or are closer replicas respectively, yet still capture the properties of the original.

\label{sec:generation}
\begin{table*}[ht]
	\centering 
	\caption{Statistics of \textsc{Cora-ML} and the graphs generated by \name and the baselines, averaged over 5 trials.
	\name closely matches the input networks in most properties, while other methods either deviate significantly in at least one statistic or overfit.
	* indicates values for the conf. model that by definition exactly match the original. 
	}
	\label{tab:gen_statistics}
	\resizebox{1\textwidth}{!}{
		\setlength{\tabcolsep}{3pt}
		\tiny
		\begin{tabular}{l r| cccccccc | c}
			\textbf{Graph}&  & \makecell{\textbf{Max}.\\\textbf{degree}} & \makecell{\textbf{Assort-}\\\textbf{ativity}} & \makecell{\textbf{Triangle}\\\textbf{count}} & \makecell{\textbf{Power}\\\textbf{law exp.}} & \makecell{\textbf{Inter-comm.}\\\textbf{unity density}} & \makecell{\textbf{Intra-comm.}\\\textbf{unity density}} 
			& \makecell{\textbf{Cluster-}\\\textbf{ing coeff.}}  & \makecell{\textbf{Charac.}\\\textbf{path len.}} & \makecell{\textbf{Average}\\\textbf{rank}}
			\\ \hline
			
			\textsc{Cora-ML}& & 240 & -0.075 & 2,814 & 1.860                 & 4.3e{-4} & 1.7e{-3}  & 2.73e{-3} & 5.61 &    \\ \hline
			Conf. model &(1\% EO)& \ditto & -0.030 & 322 & \ditto            & 1.6e{-3} & 2.8e{-4}  & 3.00e{-4} & 4.38 & 7.50   \\
			Conf. model & (52\% EO) & \ditto & -0.051 & 626 & \ditto         & 9.8e{-4} & 9.9e{-4}  & 6.10e{-4} & 4.46 & 5.83   \\
			\SBM & \hspace{-1mm} (11\% EO) & 165 & -0.052 & 1,403 & 1.814    & 6.7e{-4} & 1.2e{-3}  & 3.30e{-3} & 5.12 & 3.36  \\
			ERGM & (56\% EO) & 243 & -0.077 & 2,293 & 1.786                  & 6.9e{-4} & 1.2e{-3}  & 2.17e{-3} & 4.59 & 2.88   \\
			BTER & (2.2\% EO) & 199 & 0.033 & 3,060 & 1.787                  & 1.0e{-3} & 7.5e{-4}  & 4.62e{-3} & 4.59 & 4.75  \\
			VGAE & (0.3\% EO) & 13 & -0.009 & 14 & 1.674                     & 1.4e{-3} & 3.2e{-4}  & 1.17e{-3} & 5.28 & 5.88  \\ \hline 		
			\name \textsc{Val} & (39\% EO) & 199 & -0.060 &  1,410 & 1.773   & 6.5e{-4} & 1.3e{-3}  & 2.33e{-3} & 5.17 & 3.00  \\
			\name \textsc{EO} & (52\% EO) & 233 & -0.066 & 1,588 & 1.793     & 6.0e{-4} & 1.4e{-3}  & 2.44e{-3} & 5.20 & 1.75  \\
		\end{tabular}
	}
\end{table*}

\subsection{Assembling the Adjacency Matrix}
\label{sec:model-adjacency}
After finishing the training, we use the generator $G$ to construct a score matrix $\bs S$ of transition counts, i.e.\ we count how often an edge appears in the set of generated random walks (typically, using a much larger number of random walks than for early stopping, e.g., 500K).
While the raw counts matrix $\bs S$ is sufficient for link prediction purposes, we need to convert it to a binary adjacency matrix $\bs{\hat{A}}$ if we wish to reason about the synthetic graph.
First, $\bs S$ is symmetrized by setting $s_{ij} = s_{ji} = \max\{s_{ij}, s_{ji}\}$.
Because we cannot explicitly control the starting node of the random walks generated by $G$, some high-degree nodes will likely be overrepresented.
Thus, a simple binarization strategy like thresholding or choosing top-$k$ entries might lead to leaving out the low-degree nodes and producing singletons.

To address this issue, we use the following approach:
(i) We ensure that every node $i$ has at least one edge by sampling a neighbor $j$ with probability $p_{ij} = \frac{s_{ij}}{\sum_{v} s_{iv}}$.
If an edge was already sampled before, we repeat the procedure;
(ii) We continue sampling edges without replacement using for each edge $(i, j)$ the probability $p_{ij} = \frac{s_{ij}}{\sum_{u, v} s_{uv}}$, until we reach the desired amount of edges (e.g., as many edges as in the original graph). To obtain an undirected graph for every edge $(i, j)$ we also include $(j, i)$. 
Note that this procedure is not guaranteed to produce a fully connected graph.

\section{Experiments}
\label{sec:experiments}

In this section we evaluate the quality of the graphs generated by \name by computing various graph statistics.
We quantify the generalization power of the proposed model by evaluating its link prediction performance.
Furthermore, we demonstrate how we can generate graphs with smoothly changing properties via latent space interpolation. Additional experiments are provided in the supp.\ mat.

\textbf{Datasets. }
For the experiments we use five well-known citation datasets and the Political Blogs dataset. 
For the large \textsc{Cora} dataset and its commonly used subset of machine learning papers denoted with \textsc{Cora-ML} 
we use the same preprocessing as in \citet{bojchevski2018deep}.
For all the experiments we treat the graphs as undirected and only consider the largest connected component (LCC).
Information about the datasets is listed in Table \ref{datasets}.
\begin{table}[h]
	\centering
	\caption{Dataset statistics. $\mathbf{N_{LCC}}, \mathbf{E_{LCC}}$ - number of nodes and edges respectively in the largest connected component.}
	\label{datasets}
	\resizebox{\linewidth}{!}{
		\begin{tabular}{l| rrl}
			\textbf{Name}                 & $\mathbf{N_{LCC}}$ & $\mathbf{E_{LCC}}$ & \textbf{Reference}\\ \hline 
			\textsc{Cora-ML}                       & 2,810  & 7,981 & \cite{mccallum2000automating}\\ 
			\textsc{Cora}                          & 18,800 & 64,529 &  \cite{mccallum2000automating}\\
			\textsc{CiteSeer}                      & 2,110  & 3,757 &\cite{sen2008collective}\\		
			\textsc{Pubmed}                        & 19,717 & 44,324 & \cite{sen2008collective}\\
			\textsc{DBLP}                          & 16,191 & 51,913 & \cite{pan2016tri}\\ 
			\textsc{Pol. Blogs}                    & 1,222 & 16,714 & \cite{adamic2005political}\\
	\end{tabular}
	}
\end{table}
\subsection{Graph Generation}
\begin{figure*}[]
	\captionsetup[subfigure]{justification=centering}
	
	\centering
	\includegraphics[width=0.64 \textwidth]{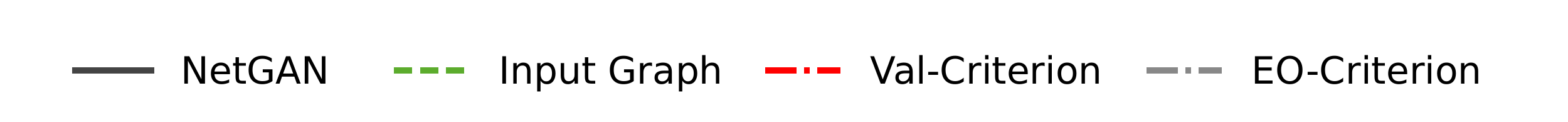}
	
	\begin{subfigure}{0.33 \textwidth}
		\includegraphics[width=\textwidth]{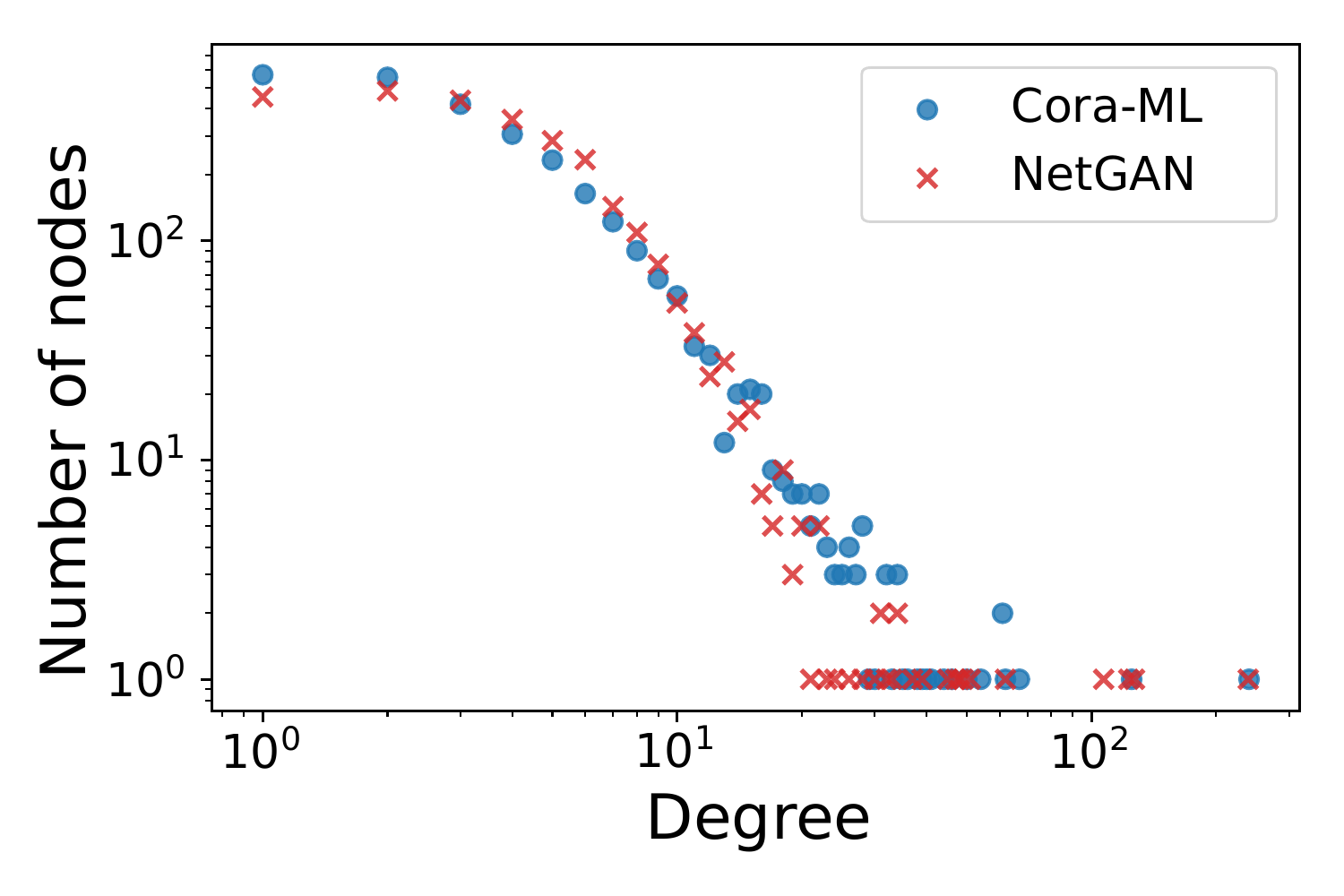}
		\caption{Degree distribution\\~}
		\label{fig:distribution_cora}
	\end{subfigure}
	\begin{subfigure}{0.33 \textwidth}
		\includegraphics[width=\textwidth]{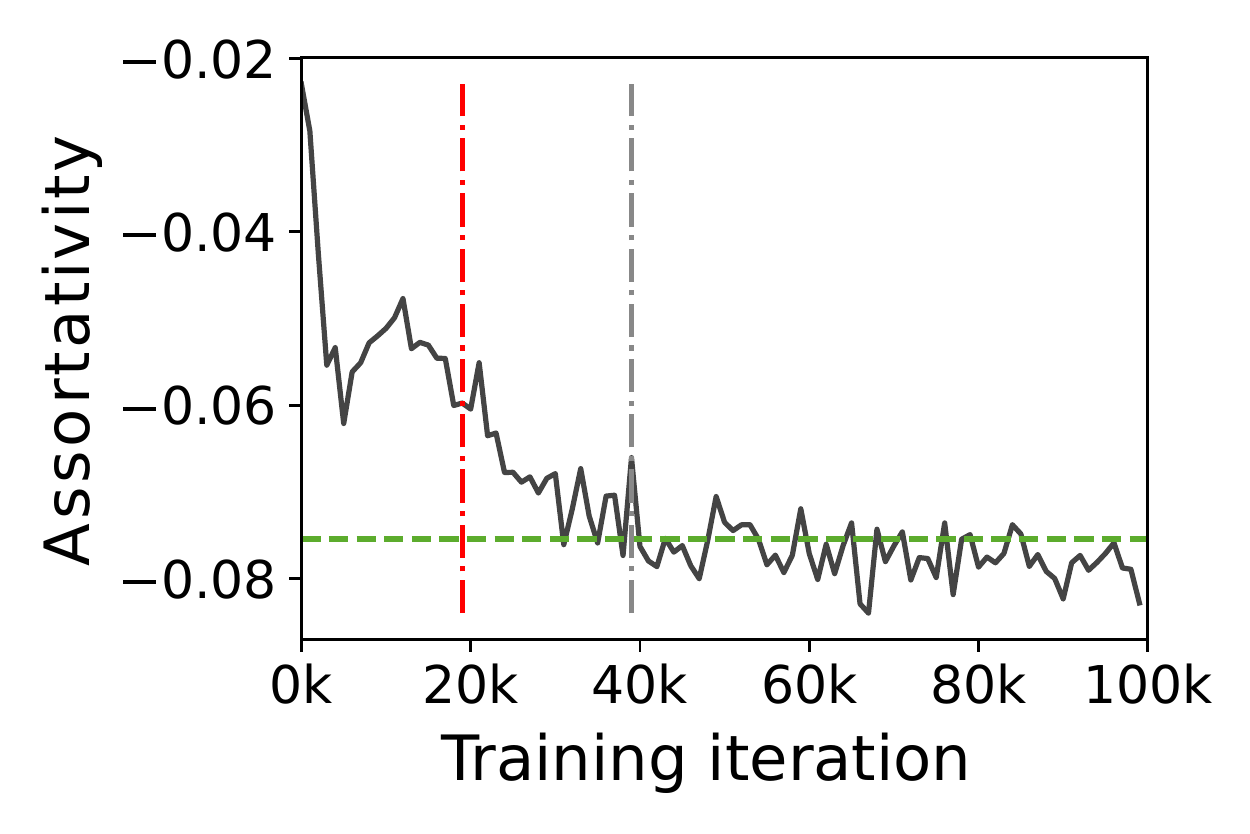}
		\caption{Assortativity over\\training iterations}
		\label{fig:cora_assortativity}
	\end{subfigure}
	\begin{subfigure}{0.33 \textwidth}
		\includegraphics[width=\textwidth]{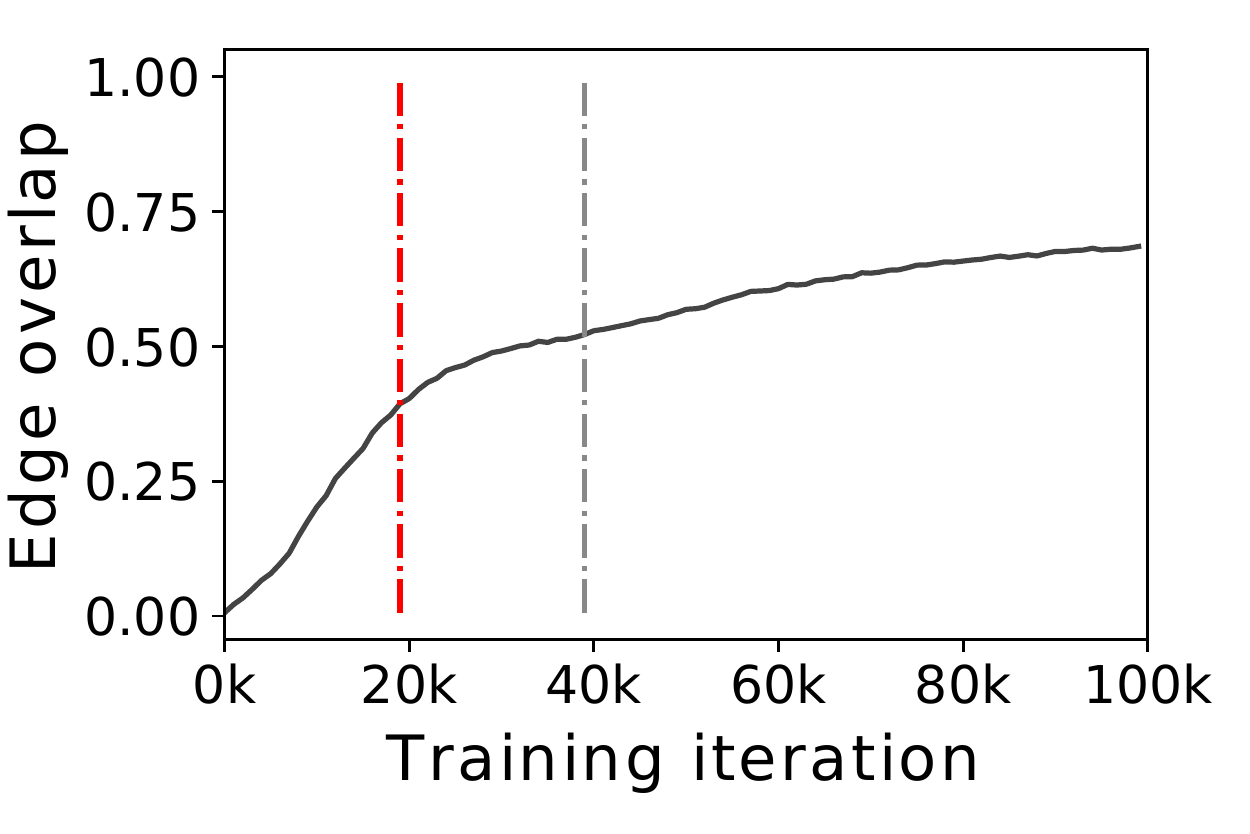}
		\caption{Edge overlap (EO) over \\training iterations}
		\label{fig:cora_edge_overlap}
	\end{subfigure}
	\caption{Properties of graphs generated by \name trained on \textsc{Cora-ML}.}
	\label{fig:degree_distribution}
\end{figure*}

\textbf{Setup. }
In this task, we fit \name to the \textsc{Cora-ML} and \textsc{Citeseer} citation networks in order to evaluate quality of the generated graphs.
We compare to the following baselines: configuration model \cite{molloy1995critical}, degree-corrected stochastic blockmodel (DC-SBM) \cite{karrer2011stochastic}, exponential random graph model (ERGM) \cite{holland1981exponential} and the block two-level Erd{\H{o}}s-R{\'e}niy random graph model (BTER) \cite{seshadhri2012bter}.
Additionally, we use the variational graph autoencoder (VGAE) \cite{kipf2016variational} as a representative of network embedding approaches.
We randomly hide $15\%$ of the edges (which are used for the stopping criterion; see Sec.~\ref{sec:training}) and fit all the models on the remaining graph.
We sample 5 graphs from each of the trained models and report their average statistics in Table \ref{tab:gen_statistics}.
Definitions of the statistics, additional metrics, standard deviations and details about the baselines are given in the supplementary material.

\textbf{Evaluation. } The general trend that becomes apparent from the results in Table \ref{tab:gen_statistics} (and Table 2 in supplementary material) is that prescribed models excel at recovering the statistics that they directly model (e.g., degree sequence for DC-SBM).
At the same time, these models struggle when dealing with graph properties that they don't account for (e.g., assortativity for BTER).
On the other hand, \name is able to capture all the graph properties well, although none of them are explicitly specified in its model definition.
We also see that VGAE is not able to produce realistic graphs.
This is expected, since the main purpose of VGAE is learning node embeddings, and not generating entire graphs.

The final column shows the average rank of each method for all statistics, with \name performing the best.
ERGM seems to be performing surprisingly well, however it suffers from  
severe overfitting -- using the same fitted ERGM for the link prediction task we get both AUC and AP scores close to 0.5 (worst possible value).
In contrast, \name does a good job both at preserving properties in generated graphs, as well as generalizing, as we see in Sec. \ref{sec:link_prediction}.

Is the good performance of \name in this experiment only due to the overlapping edges (existing in the input graph)?
To rule out this possibility we perform the following experiment:
We take the graph generated by \name, fix the overlapping edges and rewire the rest according to the configuration model.
The properties of the resulting graph (row \#3 in Table \ref{tab:gen_statistics}) deviate strongly from the input graph.
This confirms that \name does not simply memorize some edges and generates the rest at random, but rather captures the underlying structure of the network.

In line with our intuition, we can see that higher EO leads to generated graphs with statistics closer to the original.
Figs. \ref{fig:cora_assortativity} and \ref{fig:cora_edge_overlap} show how the graph statistics evolve during the training process.
Fig.~\ref{fig:cora_edge_overlap} shows that the edge overlap smoothly increasing with the number of epochs.
We provide plots for other statistics and  for \textsc{Citeseer} in the supp.\ mat.
\subsection{Link Prediction}
\label{sec:link_prediction}

\textbf{Setup. } 
Link prediction is a common graph mining task where the goal is to predict the existence of unobserved links in a given graph. 
We use it to evaluate the generalization properties of \name.
We hold out 10\% of edges from the graph for validation and 5\% as the test set, along with the same amount of randomly selected non-edges, while ensuring that the training network remains connected.
We measure the performance with two commonly used metrics: area under the ROC curve (AUC) and average precision (AP).
To evaluate \name's performance, we sample a given number of random walks (500K/100M) from the trained generator and
we use the observed transition counts between any two nodes as a measure of how likely there is an edge between them.
We compare with \SBM, node2vec and VGAE,
as well as Adamic/Adar\cite{adamic2003friends}.

\textbf{Evaluation. } The results are listed in Table \ref{tab:lp}.
There is no overall dominant method, with different methods achieving best results on different datasets. 
\name shows competitive performance for all datasets, even achieving state-of-the-art results for some of them 
(\textsc{Citeseer} and \textsc{PolBlogs}), despite not being explicitly trained for this task.

Interestingly, the \name performance increases when increasing the number of random walks sampled from the generator. 
This is especially true for the larger networks (\textsc{Cora}, \textsc{DBLP}, \textsc{Pubmed}), since given their size we need more random walks to cover the entire graph. 
This suggests that for an additional computational cost one can get significant gains in link prediction performance.
Note, that while 100M may seem like a large number, the sampling procedure can be trivially parallelized.
\begin{table*}[ht]
	\caption{Link prediction performance (in \%).}
	\resizebox{\textwidth}{!}{
		\begin{tabular}{lcccccccccccc}
			\multirow{ 2}{*}{\textbf{Method}} & \multicolumn{2}{c}{\textsc{Cora-ML}} & \multicolumn{2}{c}{\textsc{Cora}} & \multicolumn{2}{c}{\textsc{Citeseer}} & \multicolumn{2}{c}{\textsc{DBLP}} & \multicolumn{2}{c}{\textsc{Pubmed}} & \multicolumn{2}{c}{\textsc{PolBlogs}} \\
			& AUC & AP & AUC & AP & AUC & AP & AUC & AP & AUC & AP & AUC & AP  \\
			\hline
			Adamic/Adar  & 92.16 & 85.43 & 93.00 & 86.18 & 88.69 & 77.82 & 91.13 & 82.48 & 84.98 & 70.14 & 85.43 & 92.16 \\
			\SBM & \textbf{96.03} & 95.15 & 98.01 & 97.45 & 94.77 & 93.13 & \textbf{97.05} & 96.57 & \textbf{96.76} & 95.64 & 95.46 & \textbf{94.93} \\
			node2vec     & 92.19 & 91.76 & \textbf{98.52} & \textbf{98.36} & 95.29 &94.58 & 96.41 & 96.36 & 96.49 & 95.97 & 85.10 & 83.54\\
			VGAE         & 95.79 & \textbf{96.30} & 97.59 & 97.93 & 95.11 & 96.31 & 96.38 & \textbf{96.93} & 94.50 & \textbf{96.00} & 93.73 & 94.12 \\ \hline
			\name (500K) & 94.00 & 92.32 & 82.31 & 68.47 & 95.18 & 91.93 & 82.45 & 70.28 & 87.39 & 76.55 & 95.06 & 94.61 \\
			\name (100M) & 95.19 & 95.24 & 84.82 & 88.04 & \textbf{96.30} & \textbf{96.89} & 86.61 & 89.21 & 93.41 & 94.59 & \textbf{95.51} & 94.83\\
	\end{tabular}
	}
	\label{tab:lp}
\end{table*}
\begin{figure*}[h]
	\captionsetup[subfigure]{justification=centering}
	\centering
	\begin{subfigure}{0.24 \textwidth}
		\includegraphics[height=0.15 \textheight]{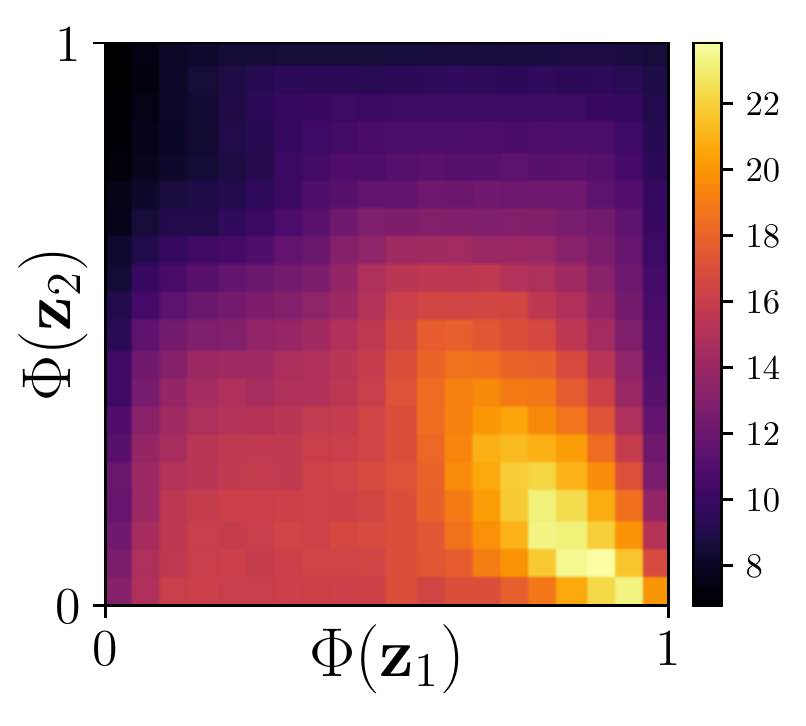}
		\caption{Avg. degree\\of start node}
		\label{z_a}
	\end{subfigure}
	\begin{subfigure}{0.24 \textwidth}
		\includegraphics[height=0.15 \textheight]{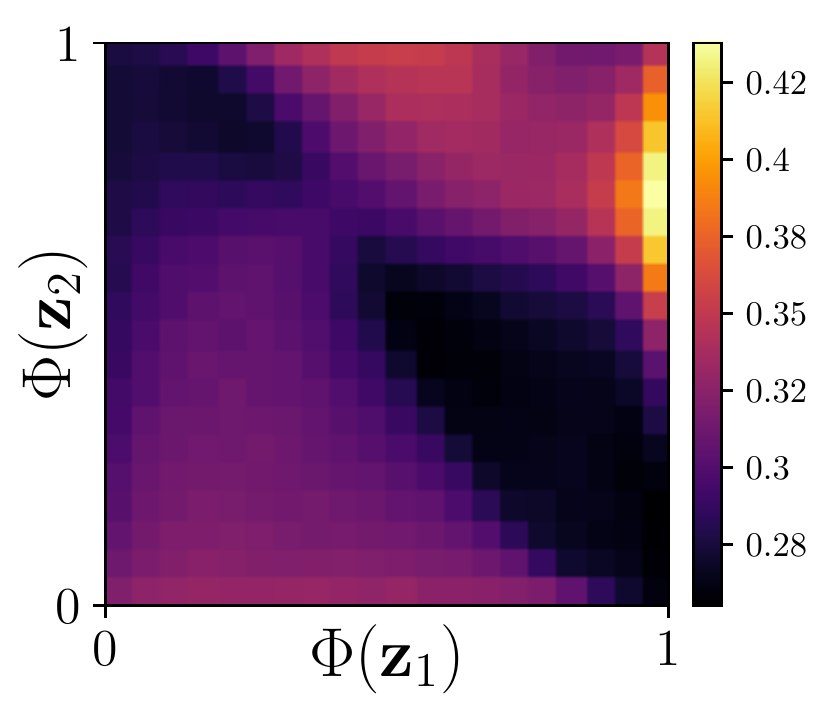}	
		\caption{Avg. share of nodes \\in start community}
		\label{z_b}
	\end{subfigure}
	\begin{subfigure}{0.24 \textwidth}
		\includegraphics[height=0.15 \textheight]{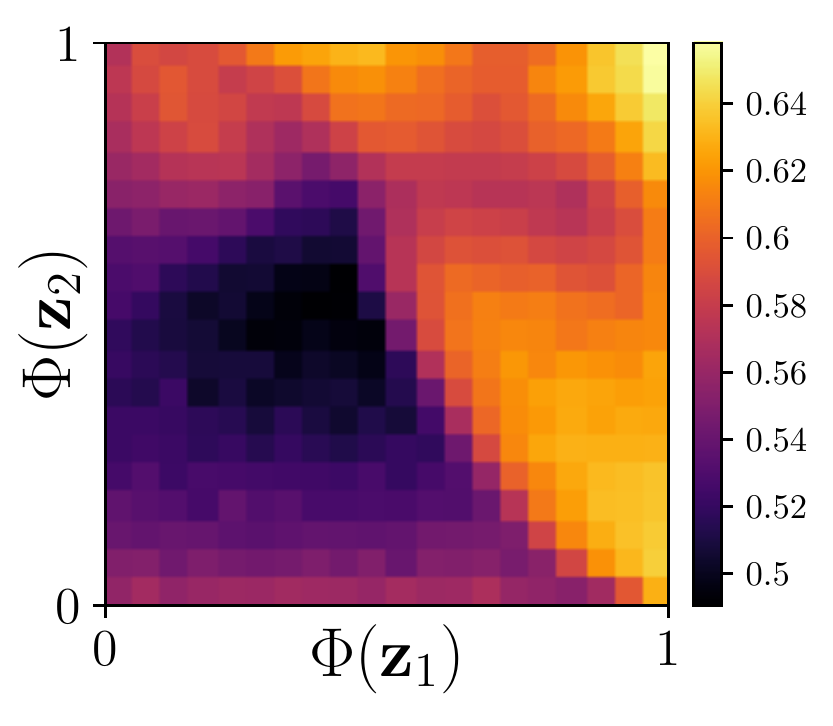}
		\caption{Gini coefficient \\(input graph: 0.48)}	
		\label{z_c}
	\end{subfigure}
	\begin{subfigure}{0.24 \textwidth}
		\includegraphics[height=0.15 \textheight]{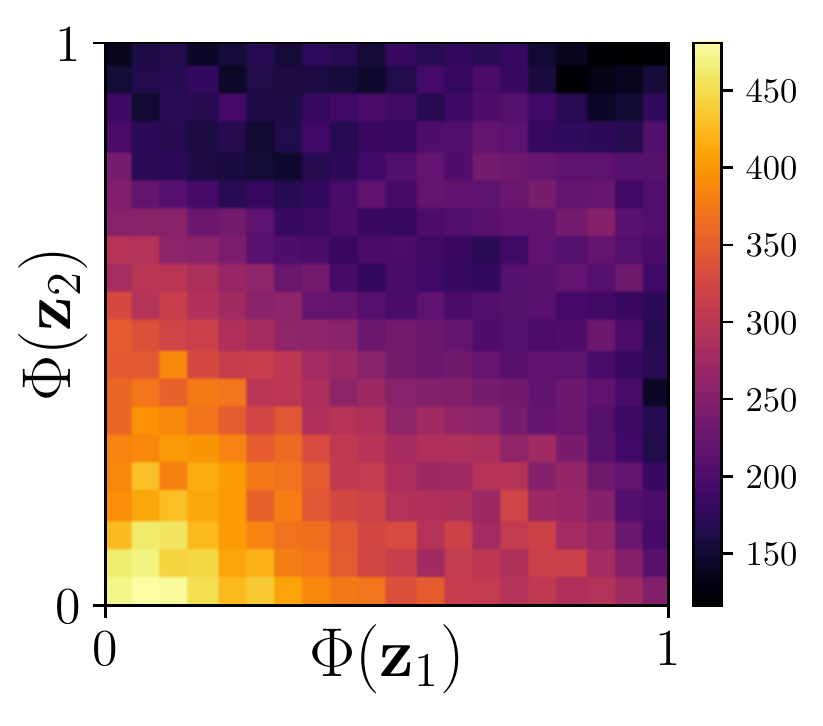}	
		\caption{Max. degree \\(input graph: 240)}
		\label{z_d}
	\end{subfigure}
	
	\caption{Properties of the random walks (\ref{z_a} and \ref{z_b}) as well as the graphs (\ref{z_c} and \ref{z_d}) sampled from the $20 \times 20$  bins.}
	\label{fig:rw_props}
	
	\captionsetup[subfigure]{justification=centering}
	
	\centering 
	\begin{subfigure}{0.535\textwidth}
		\begin{tikzpicture}
		\node[anchor=south west,inner sep=0] (image) at (0,0) {\includegraphics[width= \textwidth]{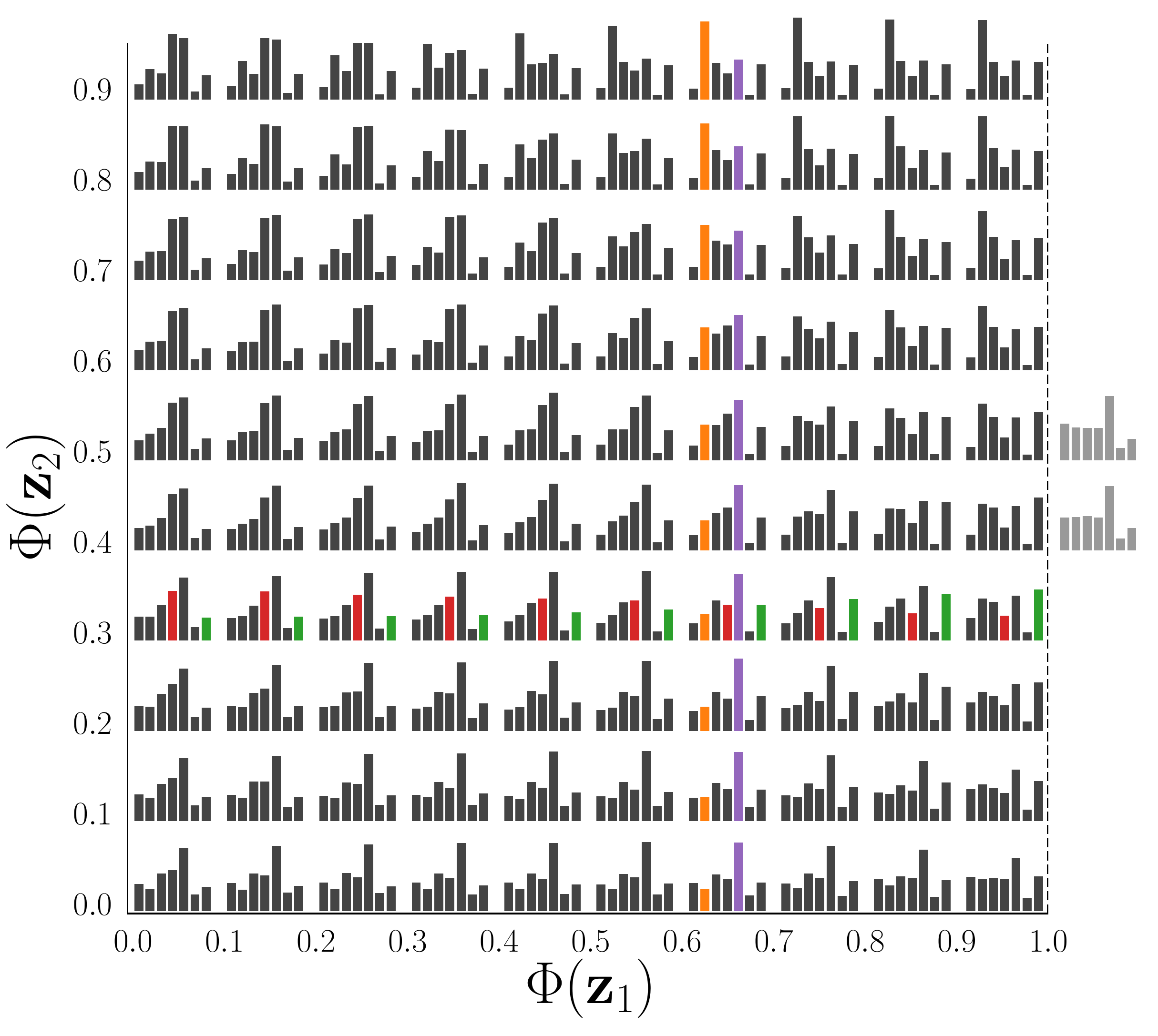}};
		\begin{scope}[x={(image.south east)},y={(image.north west)}] 
		\node[align=center] at (0.948, 0.43) {($\Omega$)};         
		\node[align=center] at (0.948, 0.65) {(*)};         
		\end{scope}
		\end{tikzpicture}
		\caption{Community histograms}
		\label{community_histograms}
	\end{subfigure}
	\hspace{10mm}
	\begin{subfigure}{0.30  \textwidth}
		\begin{subfigure}{\textwidth}
			\centering
			\includegraphics[width=1.1\textwidth]{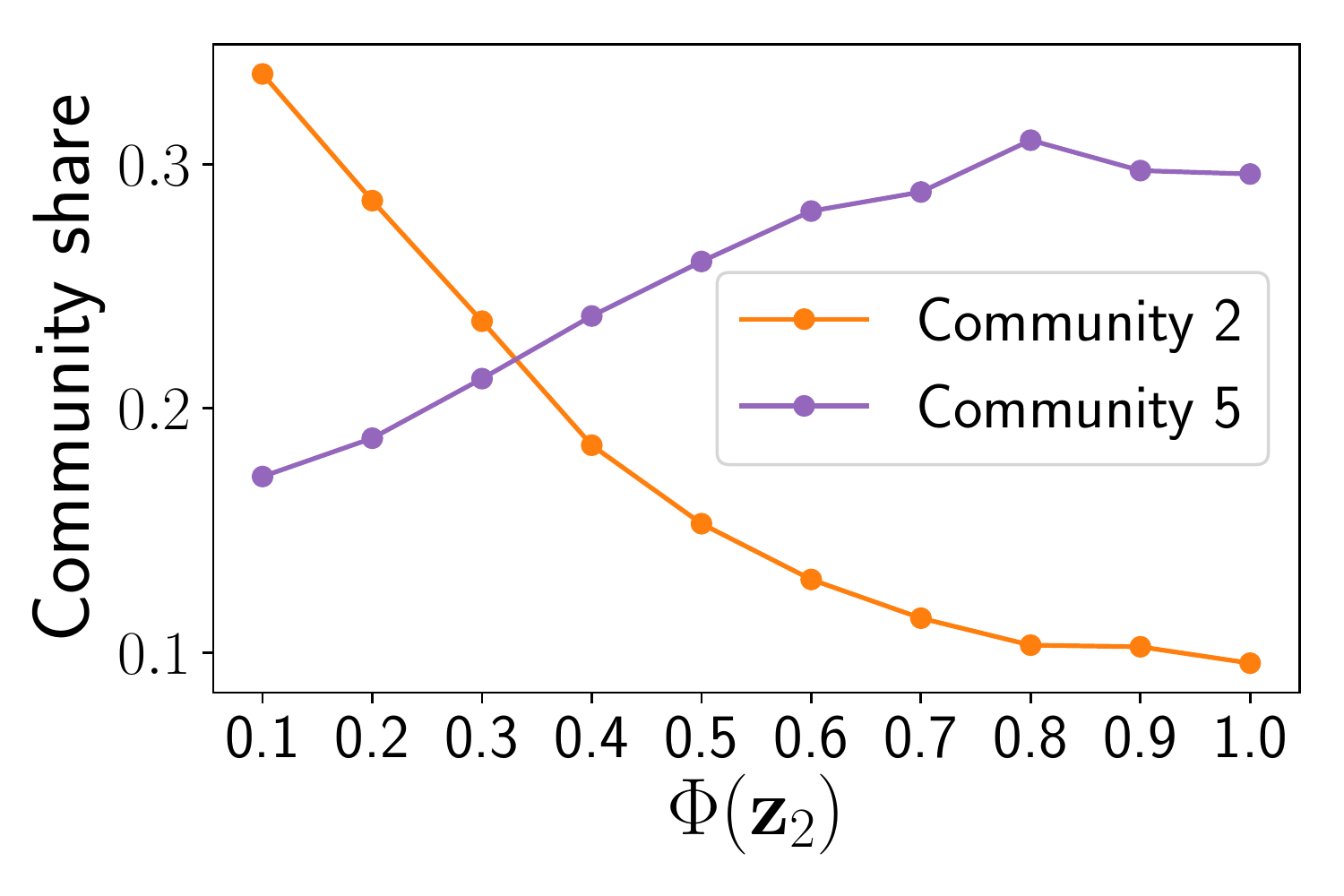}
			\caption{Top to bottom trajectory\hspace*{-13mm}}
			\label{community_distr_trajectory_1}
		\end{subfigure} 
		
		\begin{subfigure}{\textwidth}
			\centering
			\includegraphics[width=1.1\textwidth]{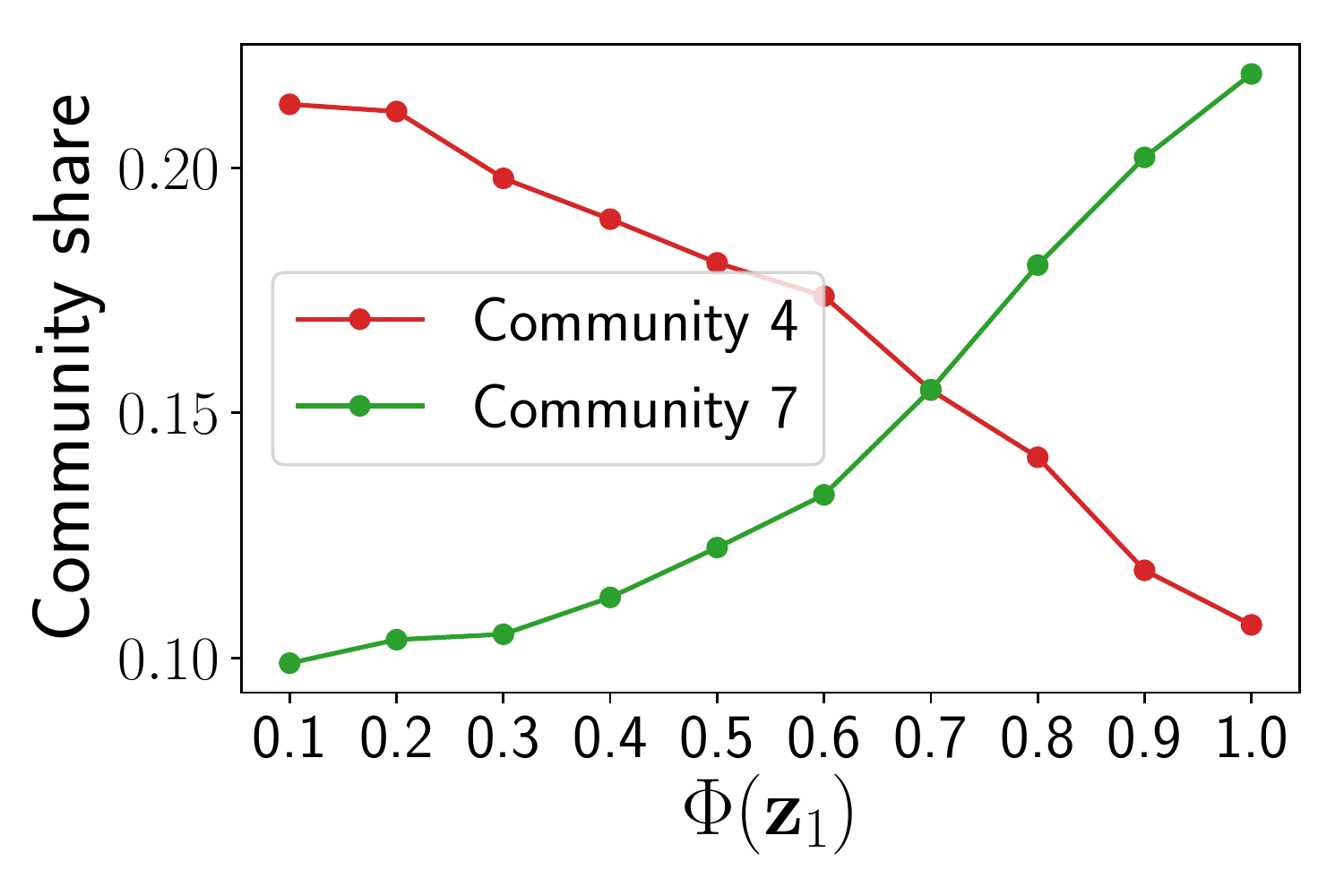}
			\caption{Left to right trajectory\hspace*{-13mm}}
			\label{community_distr_trajectory_2}
		\end{subfigure}		
	\end{subfigure}
	\caption{Community histograms of graphs sampled from subsets of the latent space. (a) shows complete community histograms on a $10\times 10$ grid. (b) and (c) show how shares of specific communities change along trajectories. ($\Omega$) is the community distribution when sampling from the entire latent space, and (*) is the community histogram of \textsc{Cora-ML}. Available as an animation at \url{https://goo.gl/bkNcVa}.}
	\label{fig:community_distr}
\end{figure*}

\textbf{Sensitivity analysis. } 
Although \name has many hyperparameters -- typical for a GAN model -- 
in practice most of them are not critical for performance, as long as they are within a reasonable range (e.g. $H\ge30$). 
\begin{wrapfigure}[12]{r}{4.75cm}
	\includegraphics[width=4.5cm]{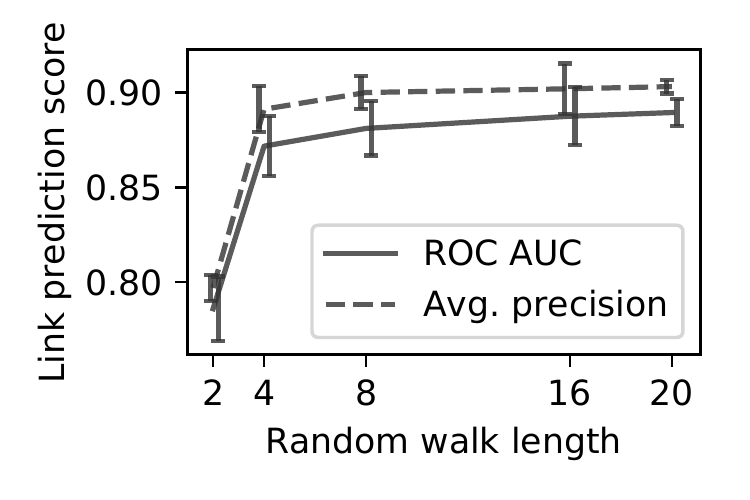}
	\caption{Effect of the random walk length $T$ on the performance.}
	\label{fig:app_rwlen_sensitivity}
\end{wrapfigure}
One important exception is the the random walk length $T$.
To choose the optimal value, we evaluate the change in link prediction performance as we vary $T$ on \textsc{Cora-ML}.
We train multiple models with different random walk lengths, and evaluate the scores ensuring each one observes equal number of transitions.
Results averaged over 5 runs are given in Fig. \ref{fig:app_rwlen_sensitivity}.
We empirically confirm that the model benefits from using longer random walks as opposed to just edges (i.e. $T$=2).
The performance gain for $T=20$ over $T=16$ is marginal and does not outweigh the additional computational cost, thus we set $T=16$ for all experiments.

\subsection{Latent Variable Interpolation}

\textbf{Setup. }
Latent space interpolation is a good way to gain insight into what kind of structure the generator was able to capture. 
To be able to visualize the properties of the generated graphs we train our model using a 2-dimensional noise vector $\bs z$ drawn as before from a bivariate standard normal distribution.
This corresponds to a 2-dimensional latent space $\Omega = \mathbb{R}^2$.
Then, instead of sampling $\bs z$ from the entire latent space $\Omega$, we now sample from subregions of $\Omega$ and visualize the results. 
Specifically, we divide $\Omega$ into $20 \times 20$ subregions (bins) of equal probability mass using the standard normal cumulative distribution function $\Phi$.
For each bin we generate 62.5K random walks. 
We evaluate properties of both the generated \emph{random walks} themselves, as well as properties of the resulting \emph{graphs} obtained by sampling a binary adjacency matrix for each bin.

\textbf{Evaluation. }
In Fig.~\ref{z_a} and \ref{z_b} we see properties of the generated random walks; in Fig.~\ref{z_c} and \ref{z_d}, we visualize properties of graphs sampled from the random walks in the respective bins.
In all four heatmaps, we see distinct patterns, e.g. higher average degree of starting nodes for the bottom right region of Fig.~\ref{z_a}, or higher degree distribution inequality in the top-right area of Fig.~\ref{z_c}. 
While Fig.~\ref{z_c} and \ref{z_d} show that certain regions of $\bs z$ correspond to generated graphs with very different degree distributions, recall that sampling from the entire latent space ($\Omega$) yields graphs with degree distribution similar to the original graph (see Fig. \ref{sibling_c}).
The model was trained on \textsc{Cora-ML}. 
More heatmaps for other metrics (16 in total) and visualizations for \textsc{Citeseer} can be found in the supplementary material.

This experiment clearly demonstrates that by interpolating in the latent space we can obtain graphs with smoothly changing properties. 
The smooth transitions in the heatmaps provide evidence that our model learns to map specific parts of the latent space to specific properties of the graph.

We can also see this mapping from latent space to the generated graph properties in the community distribution histograms on a $10\times 10$ grid in Fig.~\ref{fig:community_distr}. 
Marked by (*) and ($\Omega$) we see the community distributions for the input graph and the graph obtained by sampling on the complete latent space respectively.
In Fig.~\ref{community_distr_trajectory_1} and \ref{community_distr_trajectory_2}, we see the evolution of selected community shares when following a trajectory from top to bottom, and left to right, respectively. 
The community histograms resulting from sampling random walks from opposing regions of the latent space are very different; 
again the transitions between these histograms are smooth, as can be seen in the trajectories in Fig.~\ref{community_distr_trajectory_1}~and~\ref{community_distr_trajectory_2}.
\section{Discussion and Future Work}
When evaluating different graph generative models in Sec. \ref{sec:generation}, we observed a major limitation of explicit models.
While the prescribed approaches excel at recovering the properties directly included in their definition, they perform significantly worse with respect to the rest.
This clearly indicates the need for implicit graph generators such as \name.
Indeed, we notice that our model is able to consistently capture all the important graph characteristics (see Table \ref{tab:gen_statistics}).
Moreover, \name generalizes beyond the input graph, as can be seen by its strong link prediction performance in Sec. \ref{sec:link_prediction}.
Still, being the first model of its kind, \name possesses certain limitations, and a number of related questions could be addressed in follow-up works:

\textbf{Scalability. }
We have observed in Sec. \ref{sec:link_prediction} that it takes a large number of generated random walks to get representative transition counts for large graphs.
While sampling random walks from \name is trivially parallelizable, a possible extension of our model is to use a \emph{conditional} generator, i.e. the generator can be provided a desired starting node, thus ensuring a more even coverage.
On the other hand, the sampling procedure itself can be sped up by incorporating a hierarchical softmax output layer - a method commonly used in natural language processing.

\textbf{Evaluation. }
It is nearly impossible to judge whether a graph is realistic by visually inspecting it (unlike images, for example).
In this work we already quantitatively evaluate the performance of \name on a large number of standard graph statistics.
However, developing new measures applicable to (implicit) graph generative models will deepen our understanding of their behavior, and is an important direction for future work.

\textbf{Experimental scope. }
In the current work we focus on the setting of a single large graph.
Adaptation to other scenarios, such as a collection of smaller i.i.d. graphs, that frequently occur in other fields (e.g., chemistry, biology), would be an important extension of our model.
Studying the influence of the graph topology (e.g., sparsity, diameter) on \name's performance will shed more light on the model's properties.

\textbf{Other types of graphs. }
While plain graphs are ubiquitous, many of important applications deal with attributed, k-partite or heterogeneous networks.
Adapting the \name model to handle these other modalities of the data is a promising direction for future research.
Especially important would be an adaptation to the dynamic / inductive setting, where new nodes are added over time.
\section{Conclusion}
In this work we introduce \name - an implicit generative model for network data.
\name is able to generate graphs that capture important topological properties of complex networks, such as community structure and degree distribution, without having to manually specify any of them.
Moreover, our proposed model shows strong generalization properties, as highlighted by its competitive link prediction performance on a number of datasets.
\name can also be used for generating graphs with continuously varying characteristics using latent space interpolation.
Combined our results provide strong evidence that implicit generative models for graphs are well-suited for capturing the complex nature of real-world networks.
\section*{Acknowledgments}
This research was supported by the German Research Foundation, Emmy Noether grant GU 1409/2-1, and by the Technical University of Munich - Institute for Advanced Study, funded by the German Excellence Initiative and the European Union Seventh Framework Programme under grant agreement no 291763, co-funded by the European Union.

\bibliography{bibliography}
\bibliographystyle{icml2018}

\clearpage
\begin{table*}[ht!]
\appendix
\section{Graph statistics}
\centering
\caption{Graph statistics used to measure graph properties in this work.}
\label{tab:metrics}
\resizebox{\textwidth}{!}{
	\def\tabularxcolumn#1{m{#1}}
	\begin{tabularx}{1.2\textwidth}{Sl Sl X}
		\textbf{Metric name} & \textbf{Computation} & \textbf{Description} \\ \hline
		Maximum degree & $\underset{v\in V}{\max}\: d(v)$ & Maximum degree of all nodes in a graph. \\ 
		Assortativity & $\rho = \frac{\cov(X,Y)}{\sigma_X \sigma_Y}$ & Pearson correlation of degrees of connected nodes, where the  $(x_i,y_i)$ pairs are the degrees of connected nodes. \\
		Triangle count & $\frac{| \{ \{ u,v,w \} | \{(u,v), (v, w), (u, w)\} \subseteq E \} |}{6}$ & Number of triangles in the graph, where $u \sim v$ denotes that $u$ and $v$ are connected.\\ 		
		Power law exponent & $ 1 + n\left ( \underset{u\in V}{\sum} \log \frac{d(u)}{d_{\min}} \right )^{-1}$ & Exponent of the power law distribution, where $d_{min}$ denotes the minimum degree in a network.\\ 
		Inter-community density & $\frac{1}{K} \sum_{j=1}^K \sum_{\substack{k=1\\ k \neq j}}^K \frac{1}{\binom{|C_k|}{|C_j|}} \sum_{u \in C_j} \sum_{v \in C_k} A_{uv}$ & Fraction of possible inter-community edges present in graph.\\
		Intra-community density & $\frac{1}{K} \sum_{j=1}^K \frac{1}{\binom{|C_j|}{2}} \sum_{u, v \in C_j} A_{uv} $ & Fraction of possible intra-community edges present in graph.\\
		Wedge count & $\sum_{v \in V} {{d(v)} \choose 2}$ & Number of wedges (2-stars), i.e. two-hop paths in an undirected graph. \\ 
		Rel. edge distr. entropy & $\frac{1}{\ln |V|} \sum_{v\in V} - \frac{d(v)}{|E|} \ln \frac{d(v)}{|E|}$ & Entropy of degree distribution, 1 means uniform, 0 means a single node is connected to all others. \\ 
		LCC & $N_{max} = \underset{f\subseteq F}{\max}\: |f|$ & Size of largest connected component, where $F$  are all connected components of the graph. \\ 
		Claw count & $\sum_{v \in V} \binom{d(v)}{3}$ & Number of claws (3-stars)\\
		Gini coefficient & $\frac{2 \sum_{i=1}^{|V|} i\hat{d}_i}{|V|\sum_{i=1}^{|V|}\hat{d}_i}-\frac{|V|+1}{|V|}$ & Common measure for inequality in a distribution, where $\boldsymbol{\hat{d}}$ is the sorted list of degrees in the graph. \\ 
		Community distribution & $c_i = \frac{\sum_{v \in C_i} d(v)}{\sum_{v \in V} d(v)} $ & Share of in- and outgoing edges of community $C_i$, normalized by the number of edges in the graph. \\ 
	\end{tabularx}}
\end{table*}

\begin{figure*}
\section{Baselines}
\begin{itemize}
\item \textbf{Configuration model. } In addition to randomly rewiring \emph{all} edges in the input graph, we also generate random graphs with similar overlap as graphs generated by \name using the configuration model. For this, we randomly select a share of edges (e.g. 39\%) and keep them fixed, and shuffle the remaining edges. This leads to a graph with the specified edge overlap; in Table 2 we show that with the same edge overlap, \name's generated graphs in general match the input graph better w.r.t the statistics we measure.

\item \textbf{Exponential random graph model. } We use the R implementation of ERGM from the \texttt{ergm} package \cite{hunter2008ergm}.
We used the following parameter settings: \texttt{edge count, density, degree correlation, deg1.5, and gwesp}. Here, \texttt{deg1.5} is the sum of all degrees to the power of 1.5, and \texttt{gwesp} refers to the geometrically weighted edgewise shared partner distribution.

\item \textbf{Degree-corrected stochastic blockmodel.} We use the Python implementation from the \texttt{graph-tool} package \cite{peixoto_graph-tool_2014} using the recommended hyperparameter settings.

\item \textbf{Variational graph autoencoder. } We use the implementation provided by the authors (\url{https://github.com/tkipf/gae}). 
We construct the graph from the predicted edge probabilities using the same protocol as in Sec. 3.3 of our paper.
To ensure a fair comparison we perform early stopping, i.e. select the weights that achieve the best validation set performance.
\end{itemize}
\end{figure*}
\clearpage

\begin{table*}[h!]
\section{Properties of generated graphs}
\centering
\caption{Comparison of graph statistics between the \textsc{Citeseer/Cora-ML} graph and graphs generated by GraphGAN and DC-SBM, averaged after 5 trials.
}
\label{tab:app_statistics_citeseer}
\resizebox{\textwidth}{!}{
	\begin{tabular}{l r | l @{\hspace{0.05cm}} l l @{\hspace{0.05cm}} l l @{\hspace{0.05cm}} l l @{\hspace{0.05cm}} l l @{\hspace{0.05cm}} l l @{\hspace{0.05cm}} l l@{\hspace{0.05cm}} l}
		\textbf{Graph} & & \multicolumn{2}{c}{\makecell{\textbf{Max}.\\ \textbf{degree}}} & \multicolumn{2}{c}{\makecell{\textbf{Assortativity}}} & \multicolumn{2}{c}{\makecell{\textbf{Triangle}\\\textbf{count}}} & \multicolumn{2}{c}{\makecell{\textbf{Power law}\\\textbf{exponent}}} & \multicolumn{2}{c}{\makecell{\textbf{Avg. Inter-com-}\\\textbf{munity density}}} & \multicolumn{2}{c}{\makecell{\textbf{Avg. Intra-com-}\\\textbf{munity density}}}
		 & \multicolumn{2}{c}{\makecell{\textbf{Clustering}\\\textbf{coefficient}}}
		  \\
	& & \multicolumn{1}{c}{Avg.} & \multicolumn{1}{c}{Std.} & \multicolumn{1}{c}{Avg.} & \multicolumn{1}{c}{Std.} & \multicolumn{1}{c}{Avg.} & \multicolumn{1}{c}{Std.} & \multicolumn{1}{c}{Avg.} & \multicolumn{1}{c}{Std.} & \multicolumn{1}{c}{Avg.} & \multicolumn{1}{c}{Std.} & \multicolumn{1}{c}{Avg.} & \multicolumn{1}{c}{Std.}  & \multicolumn{1}{c}{Avg.} & \multicolumn{1}{c}{Std.} \\
	\hline

\textsc{Citeseer} & & 77 & & -0.022 & & 451 &  & 2.239 &  &   4.9e{-4} & & 9.3e{-4} & & 1.08e{-2}\\

Conf. model &  & \ditto & \ditto & -0.017 & $\pm$ 0.006 & 20 & $\pm$ 6.50 &\ditto & \ditto & 1.1e{-3} & $\pm$ 1e{-5} &  2.3e{-4} & $\pm$ 2e{-5} & 5.80e{-4} & $\pm$ 1.29e{-4} \\
Conf. model & (42\% EO)  & \ditto & \ditto & -0.020 & $\pm$ 0.009 & 54 & $\pm$ 8.8 & \ditto & \ditto & 8.4e{-4} &  $\pm$ 1e{-5} & 5.1e{-4}& $\pm$ 1e{-5} & 1.33e{-3} & $\pm$ 6.15e{-5}\\
Conf. model &  (76\% EO) & \ditto & \ditto & -0.024 & $\pm$ 0.006 & 207 & $\pm$ 11.8 & \ditto & \ditto & 6.3e{-4} & $\pm$ 1e{-5} &   7.6e{-4}& $\pm$ 1e{-5} & 5.00e{-3} & $\pm$ 2.57e{-4} \\

\SBM & (6.6\% EO) & 53 & $\pm$ 5.6 & 0.022 & $\pm$ 0.018 & 257 & $\pm$ 30.9 & 2.066 & $\pm$ 0.014 & 7.6e{-4} & $\pm$ 2e{-5} & 5.3e{-4}  & $\pm$ 3e{-5} & 1.00e{-2} & $\pm$ 2.63e{-3}\\
ERGM & (27\% EO) & 66 & $\pm$ 1 & 0.052 & $\pm$ 0.005 & 415.6 & $\pm$ 8 & 2.0 & $\pm$ 0.01 & 9.3e{-4} & $\pm$ 2e{-5} & 4.8e{-4} &$\pm$ 6e{-6} & 1.49e{-2} & $\pm$ 5.68e{-4}\\
BTER & (2\% EO)  & 70  &  $\pm$ 7.2 & 0.065 &  $\pm$ 0.014 & 449 &  $\pm$  33 &  2.049 &  $\pm$ 0.01 & 1.1e{-3} &  $\pm$ 2e{-5} & 2.8e{-4} &  $\pm$ 6e{-6} & 1.22e{-2} & $\pm$ 2.31e{-3} \\
VGAe& (0.2\% EO) &  9.2 &  $\pm$ 0.7 & -0.057 &  $\pm$ 0.016 & 2 &  $\pm$  1 & 2.039 &  $\pm$ 0.00 & 1.2e{-3} &  $\pm$ 1e{-5} & 2.5e{-4} &  $\pm$ 2e{-5}  & 1.35e{-3} & $\pm$ 9.96e{-4}\\  

\name\textsc{Val}& (42\% EO)  & 54 & $\pm$ 4.2 & -0.082 & $\pm$ 0.009 &   316 & $\pm$ 11.2  &  2.154 & $\pm$ 0.003 & 6.5e{-4} & $\pm$ 2e{-5} & 8.0e{-4} & $\pm$ 2e{-5} & 1.99e{-2} & $\pm$ 3.48e{-3} \\
\name\textsc{EO}& (76\% EO)  & 63 & $\pm$ 4.3 & -0.054 & $\pm$ 0.006 &   227 & $\pm$ 13.3  &  2.204 & $\pm$ 0.003 & 5.9e{-4} & $\pm$ 2e{-5} & 8.6e{-4} & $\pm$ 1e{-5}  & 7.71e{-3}  & $\pm$ 2.43e{-4} \\

\hline

\textsc{Cora-ML} &  & 240 & & -0.075 & & 2,814 &  & 1.86 &  & 4.3e{-4} &  & 1.7e{-3} & & 2.73e{-3} \\

Conf. model &  & \ditto & \ditto & -0.030 & $\pm$ 0.003 & 322 & $\pm$ 31 & \ditto & \ditto  & 1.6e{-3} & $\pm$ 1e{-5} & 2.8e{-4} & $\pm$ 1e{-5} & 3.00e{-4} & $\pm$ 2.88{e-5} \\
Conf. model &  (39\% EO) & \ditto & \ditto & -0.050 & $\pm$ 0.005 & 420 & $\pm$ 14 & \ditto & \ditto  & 1.1e{-3} & $\pm$ 1e{-5} & 8.0e{-4} & $\pm$ 1e{-5} & 4.10e{-4} & $\pm$ 1.40{e-5} \\
Conf. model &  (52\% EO) & \ditto & \ditto & -0.051 & $\pm$ 0.002 & 626 & $\pm$ 19 & \ditto & \ditto & 9.8e{-4} & $\pm$ 1e{-5} & 9.9e{-4} & $\pm$ 2e{-5} & 6.10e{-4} & $\pm$ 1.85{e-5} \\

\SBM & (11\% EO) & 165 & $\pm$ 9.0 & -0.052 & $\pm$ 0.004 & 1,403 & $\pm$ 67 & 1.814 & $\pm$ 0.008 & 6.7e{-4} & $\pm$ 2e{-5} & 1.2e{-3} & $\pm$ 4e{-5}  & 3.30e{-3} & $\pm$ 2.71{e-4} \\
ERGM & (56\% EO) & 243 & $\pm$ 1.94 & -0.077 & $\pm$ 0.000 & 2,293 & $\pm$ 23 & 1.786 & $\pm$ 0.003 & 6.9e{-4}  & $\pm$ 2e{-5}  & 1.2e{-3} & $\pm$ 1e{-5} & 2.17e{-3} & $\pm$ 5.44{e-5} \\
BTER & (2\% EO) & 199 &  $\pm$ 13 & 0.033 &  $\pm$ 0.008 & 3060 &  $\pm$ 114 & 1.787 &  $\pm$  0.004 & 1.1e{-3} &  $\pm$ 1e{-5} & 7.5e{-4} &  $\pm$ 1e{-5} & 4.62e{-3} & $\pm$ 5.92{e-4} \\
VGAe& (0.3\% EO) & 13.1 &  $\pm$  1 & -0.010 &  $\pm$ 0.014 & 14 &  $\pm$ 3 & 1.674 &  $\pm$ 0.001 & 1.4e{-3} &  $\pm$ 2e{-5} & 3.2e{-4}  &  $\pm$ 1e{-5}  & 1.17e{-3}   & $\pm$ 2.02{e-4} \\

\name\textsc{Val} & (39\% EO) & 199 & $\pm$ 6.7 & -0.060 & $\pm$ 0.004 &  1,410 & $\pm$ 30 &   1.773 & $\pm$ 0.002 & 6.5e{-4} & $\pm$ 1e{-5} & 1.3e{-3} & $\pm$ 2e{-5} & 2.33e{-3} & $\pm$ 1.75{e-4} \\
\name\textsc{EO} & (52\% EO) & 233 & $\pm$ 3.6 & -0.066 & $\pm$ 0.003 &  1,588 & $\pm$ 59 &   1.793 & $\pm$ 0.003 & 6.0e{-4} & $\pm$ 1e{-5} & 1.4e{-3} & $\pm$ 1e{-5} & 2.44e{-3} & $\pm$ 1.91{e-4} \\
\end{tabular}}
	
\vspace{0.3cm}
\resizebox{\textwidth}{!}{
	\begin{tabular}{l r |l @{\hspace{0.05cm}} l l @{\hspace{0.05cm}} l l @{\hspace{0.05cm}} l l @{\hspace{0.05cm}} l l @{\hspace{0.05cm}} l l @{\hspace{0.05cm}} l l@{\hspace{0.05cm}} l }
		
		\textbf{Graph} & & \multicolumn{2}{c}{\makecell{\textbf{Wedge count}}} & \multicolumn{2}{c}{\makecell{\textbf{Rel. edge}\\\textbf{distr. entr.}}} & \multicolumn{2}{c}{\makecell{\textbf{Largest}\\\textbf{conn. comp}}} & \multicolumn{2}{c}{\makecell{\textbf{Claw count}}} & \multicolumn{2}{c}{\makecell{\textbf{Gini coeff.}}} & \multicolumn{2}{c}{\makecell{\textbf{Edge overlap}}}  & \multicolumn{2}{c}{\makecell{\textbf{Characteristic}\\\textbf{path length}}} \\ 
		&  & \multicolumn{1}{c}{Avg.} & \multicolumn{1}{c}{Std.} & \multicolumn{1}{c}{Avg.} & \multicolumn{1}{c}{Std.} & \multicolumn{1}{c}{Avg.} & \multicolumn{1}{c}{Std.} & \multicolumn{1}{c}{Avg.} & \multicolumn{1}{c}{Std.} & \multicolumn{1}{c}{Avg.} & \multicolumn{1}{c}{Std.} & \multicolumn{1}{c}{Avg.} & \multicolumn{1}{c}{Std.} & \multicolumn{1}{c}{Avg.} & \multicolumn{1}{c}{Std.}\\ \hline
		
\textsc{Citeseer} & & 16,824 & & 0.959 & & 2,110 & & 125,701 & & 0.404 & &  1 & & 10.33\\

Conf. model & & \ditto & \ditto & 0.955 & $\pm$ 0.001 & 2,011 & $\pm$ 6.8 & \ditto & \ditto & \ditto & \ditto & 0.008 & $\pm$ 0.001  & 5.95 & $\pm$ 0.03\\
Conf. model & (42\% EO) & \ditto & \ditto & 0.956 & $\pm$ 0.001 & 2,045 & $\pm$ 12.5 & \ditto & \ditto & \ditto & \ditto & 0.42 & $\pm$ 0.002 & 6.14 & $\pm$ 0.03 \\
Conf. model & (76\% EO) & \ditto & \ditto & 0.957 & $\pm$ 0.001 & 2,065 & $\pm$ 10.2 & \ditto & \ditto &\ditto & \ditto & 0.76 & $\pm$ 0.0  & 6.85 & $\pm$ 0.04\\ 

\SBM & (6.6\% EO) &  15,531 & $\pm$ 592 & 0.938 & $\pm$ 0.001 & 1,697 & $\pm$ 27 & 69,818 & $\pm$ 11,969 & 0.502 & $\pm$ 0.005  & 0.066 & $\pm$ 0.011 & 7.75 & $\pm$ 0.26\\
ERGM & (27\% EO) & 16,346 & $\pm$ 101 & 0.945 & $\pm$  0.001 & 1,753 & $\pm$ 15 & 80,510 & $\pm$ 1,337 & 0.474 & $\pm$ 0.003 & 0.27 & $\pm$ 0.01 & 5.92 & $\pm$ 0.01\\ 
BTER & (2\% EO) & 18,193 &  $\pm$ 661 & 0.940 &  $\pm$ 0.001 & 1,708 &  $\pm$  14 & 113,425 &  $\pm$  19,737 & 0.491 &  $\pm$ 0.007 & 0.02 &  $\pm$  0.002 & 5.66 & $\pm$ 0.07\\
VGAe& (0.2\% EO) & 8,141 &  $\pm$ 47 & 0.986 &  $\pm$ 0.000 & 2,110 &  $\pm$ 0 & 6,611 &  $\pm$ 144 & 0.256 &  $\pm$ 0.003 & 0.002 &  $\pm$ 0.001 & 7.75 & $\pm$ 0.04\\ 

\name\textsc{Val}& (42\% EO)  &  12,998 & $\pm$ 84.6 &  0.969 & $\pm$ 0.000 &  2,079 & $\pm$ 12.6 &  57,654 &  $\pm$ 4,226 & 0.354 &  $\pm$ 0.001 & 0.42 & $\pm$ 0.006 & 8.28 & $\pm$ 0.11\\
\name\textsc{EO}& (76\% EO)  & 15,202 & $\pm$ 378 &  0.963 & $\pm$ 0.000 &  2,053 & $\pm$ 23 &  94,149 &  $\pm$ 11,926 & 0.385 &  $\pm$ 0.002 & 0.76 &$\pm$ 0.01 & 7.68 & $\pm$ 0.13\\

\hline

\textsc{Cora-ML}&  & 101,872 & & 0.941 & & 2,810 & & 3.1e6 & & 0.482 & & 1 & & 5.61\\

Conf. model & &  \ditto & \ditto & 0.928 & $\pm$ 0.002 & 2,785 & $\pm$ 4.9 &  \ditto & \ditto & \ditto & \ditto & 0.013 & $\pm$ 0.001 & 4.38 & $\pm$ 0.01\\
Conf. model & (39\% EO) & \ditto &  \ditto & 0.931 & $\pm$ 0.002 & 2,793 & $\pm$ 2.0 &  \ditto & \ditto & \ditto & \ditto & 0.39 & $\pm$ 0.0 & 4.41 & $\pm$ 0.02\\
Conf. model & (52\% EO) &\ditto & \ditto & 0.933 & $\pm$ 0.001 & 2,793 & $\pm$ 6.0 &  \ditto & \ditto & \ditto & \ditto & 0.52 & $\pm$ 0.0 & 4.46 & $\pm$ 0.02\\

\SBM & (11\% EO)  & 73,921 & $\pm$ 3,436 & 0.934 & $\pm$ 0.001 & 2,474 & $\pm$ 18.9 & 1.2e6 & $\pm$ 170,045 & 0.523 & $\pm$ 0.003 & 0.11 & $\pm$ 0.003 & 5.12 & $\pm$ 0.04 \\
ERGM & (56\% EO) & 98,615 & $\pm$ 385 & 0.932 & $\pm$ 0.001 & 2,489 & $\pm$ 11 & 3,1e6 & $\pm$ 57,092 & 0.517 & $\pm$ 0.002 & 0.56 & $\pm$ 0.014 & 4.59 & $\pm$ 0.02\\
BTER & (2\% EO)  & 91,813 &  $\pm$ 3,546 & 0.935 &  $\pm$ 0.000 & 2,439 &  $\pm$ 19 & 2.0e6 &  $\pm$ 280,945 & 0.515 &  $\pm$ 0.003 & 0.02 &  $\pm$  0.001 & 4.59 & $\pm$ 0.03\\
VGAe& (0.3\% EO) & 31,290 &  $\pm$ 178 & 0.990 &  $\pm$  0.000 & 2,810 &  $\pm$ 0 & 46,586 &  $\pm$ 937 & 0.223 &  $\pm$ 0.003 & 0.003 &  $\pm$ 0.001 & 5.28 & $\pm$ 0.01\\

\name\textsc{Val}& (39\% EO)  &  75,724 & $\pm$ 1,401 &  0.959 & $\pm$ 0.000 &  2,809 & $\pm$ 1.6 & 1.8e6 & $\pm$ 141,795 & 0.398 & $\pm$ 0.002 & 0.39 & $\pm$ 0.004 & 5.17 & $\pm$ 0.04\\
\name\textsc{EO} &  (52\% EO)  & 86,763 & $\pm$ 1,096 &  0.954 & $\pm$ 0.001 &  2,807 & $\pm$ 1.6 & 2.6e6 & $\pm$ 103,667 & 0.42 & $\pm$ 0.003 & 0.52 & $\pm$ 0.001 & 5.20 & $\pm$ 0.02\\
	
	\end{tabular}}
	
\end{table*}

\begin{figure*}
\section{Graph statistics during the training process}
	\centering
	\includegraphics[width= 0.8\textwidth]{figures/legend}
	
	\begin{subfigure}{0.32 \textwidth}
	\includegraphics[width=\textwidth]{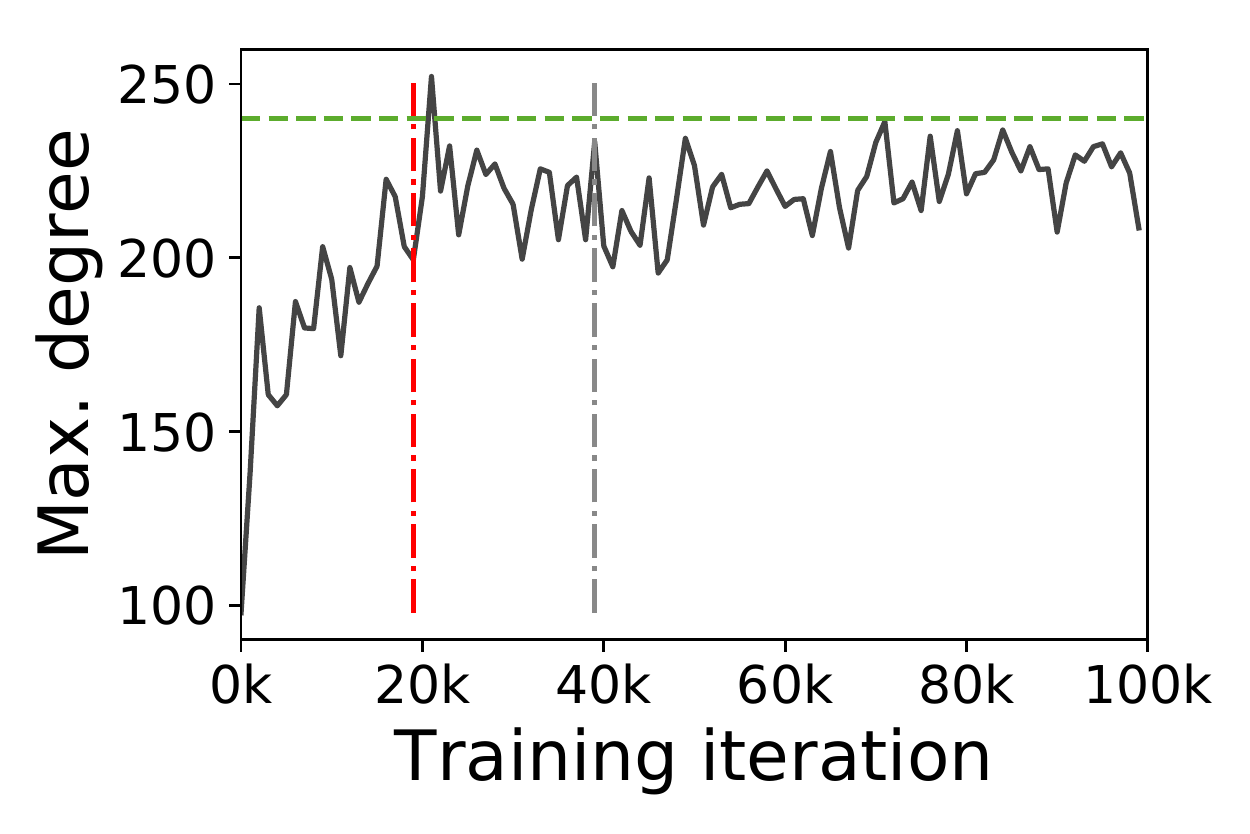}
	\caption{}
	\end{subfigure}
	\begin{subfigure}{0.32 \textwidth}
	\includegraphics[width=\textwidth]{figures/convergence_cora/assortativity}
	\caption{}
	\end{subfigure}
	\begin{subfigure}{0.32 \textwidth}
	\includegraphics[width=\textwidth]{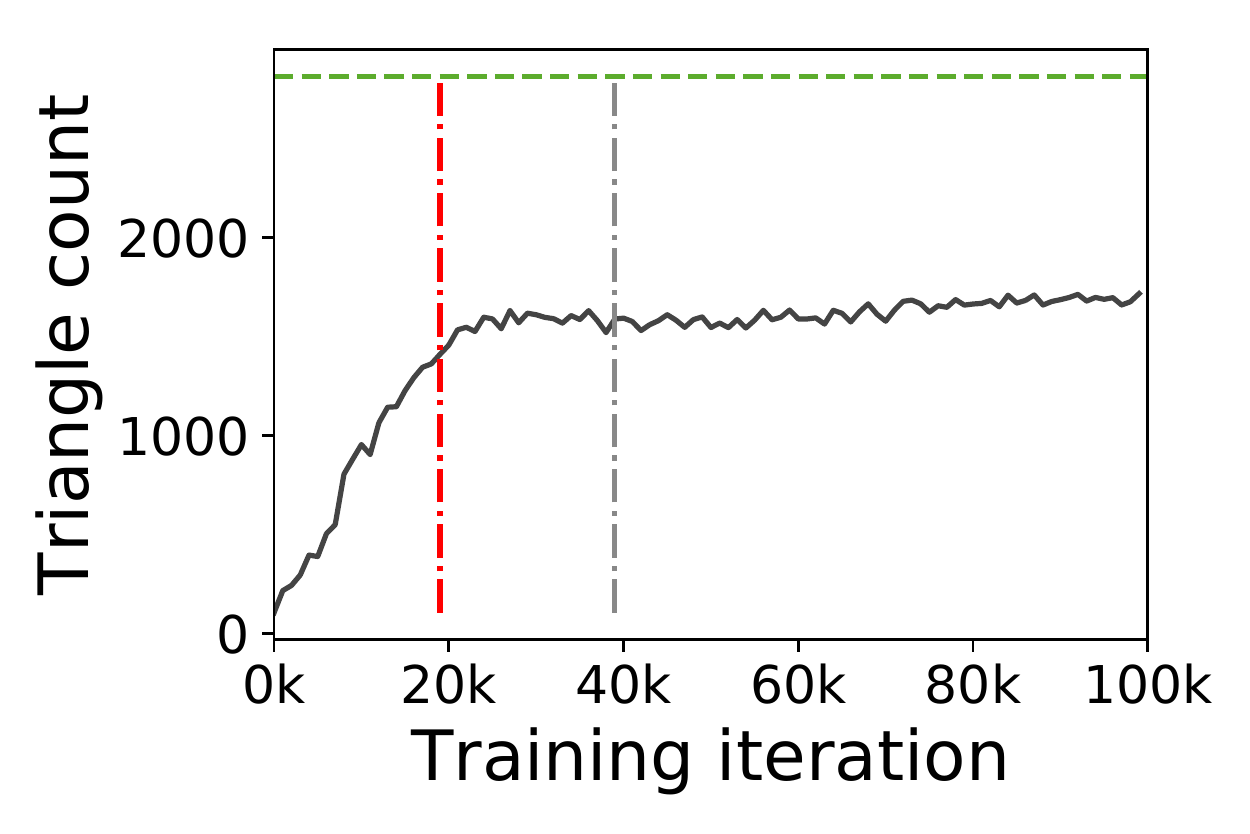}
	\caption{}
	\end{subfigure}
	
	\begin{subfigure}{0.32 \textwidth}
	\includegraphics[width=\textwidth]{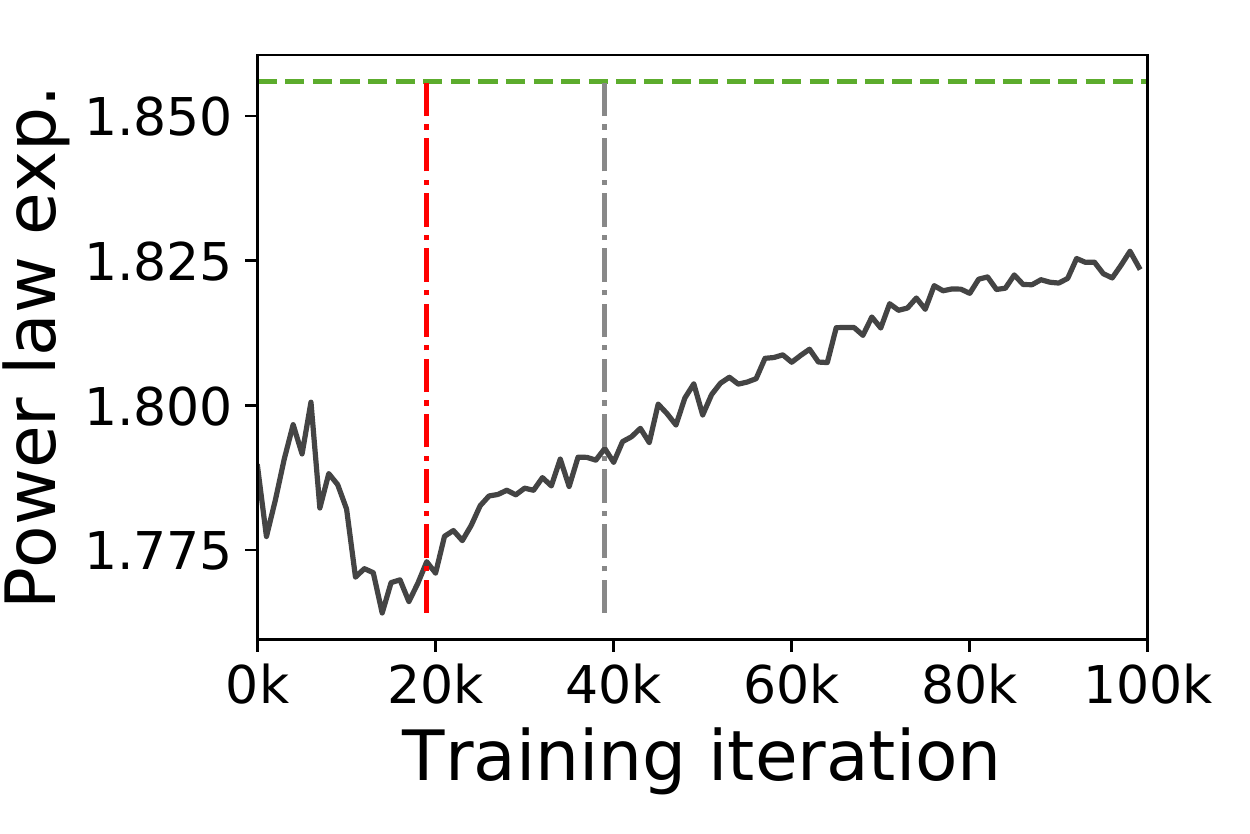}
	\caption{}
	\end{subfigure}
	\begin{subfigure}{0.32 \textwidth}
	\includegraphics[width=\textwidth]{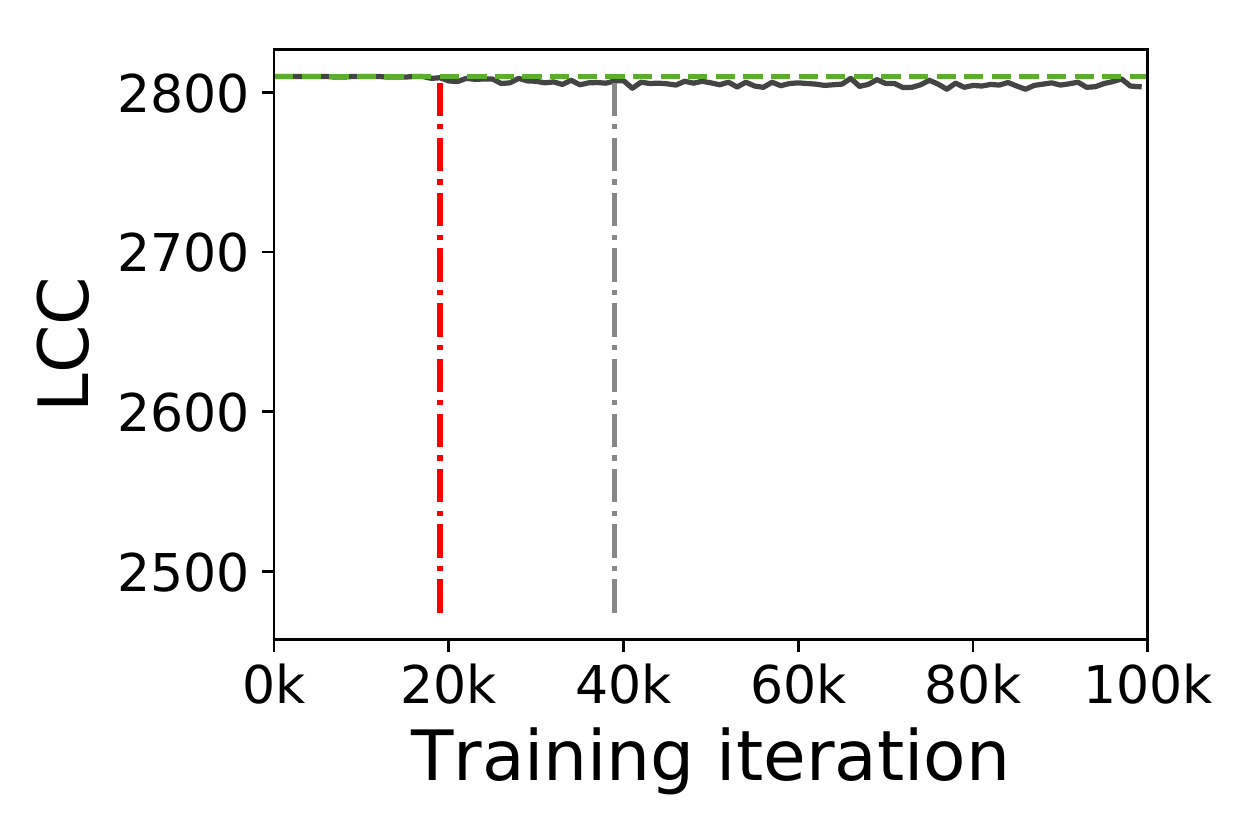}
	\caption{}
	\end{subfigure}
	\begin{subfigure}{0.32 \textwidth}
	\includegraphics[width=\textwidth]{figures/convergence_cora/overlap}
	\caption{}
	\end{subfigure}
	\caption{Evolution of graph statistics during training on \textsc{Cora-ML}}
\label{fig:app_degree_distribution_cora}
\end{figure*}

\begin{figure*}

	\centering
	\includegraphics[width= 0.8\textwidth]{figures/legend}
	
	\begin{subfigure}{0.32 \textwidth}
	\includegraphics[width=\textwidth]{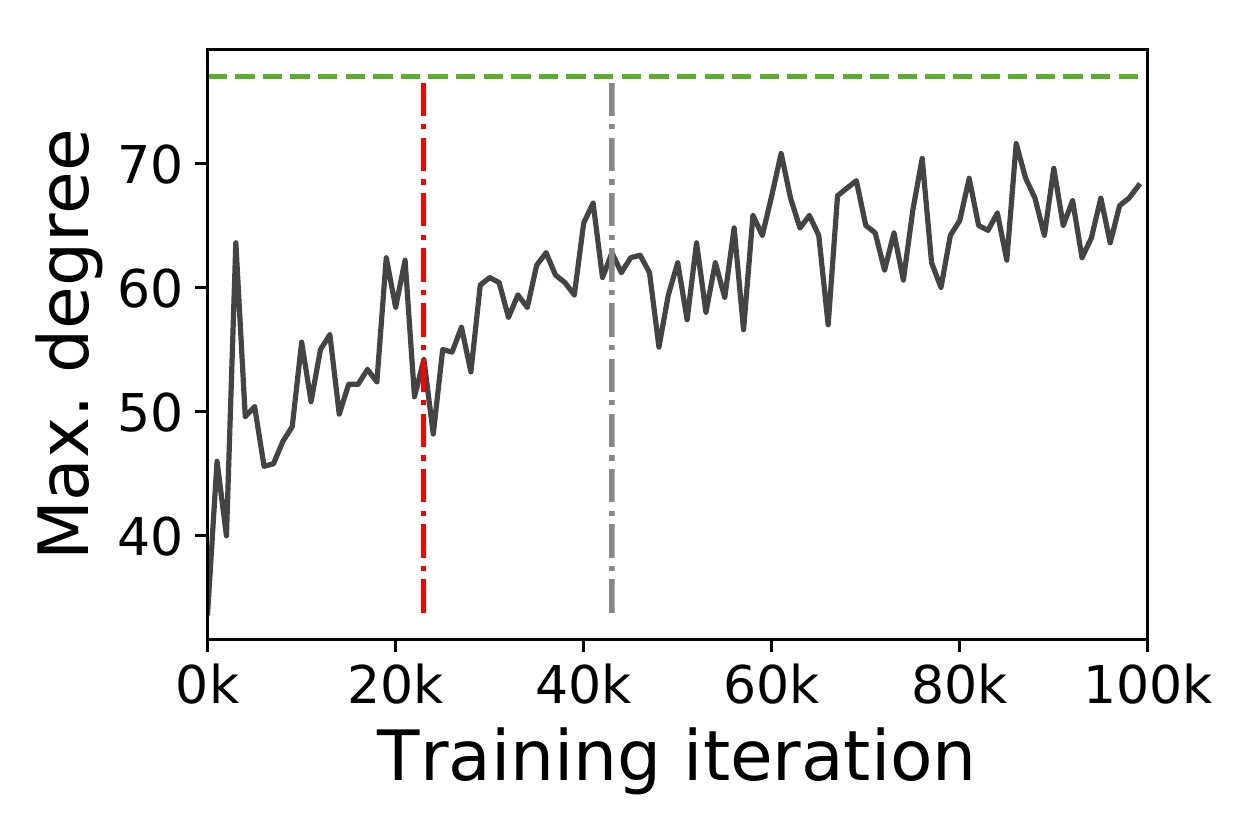}
	\caption{}
	\end{subfigure}
	\begin{subfigure}{0.32 \textwidth}
	\includegraphics[width=\textwidth]{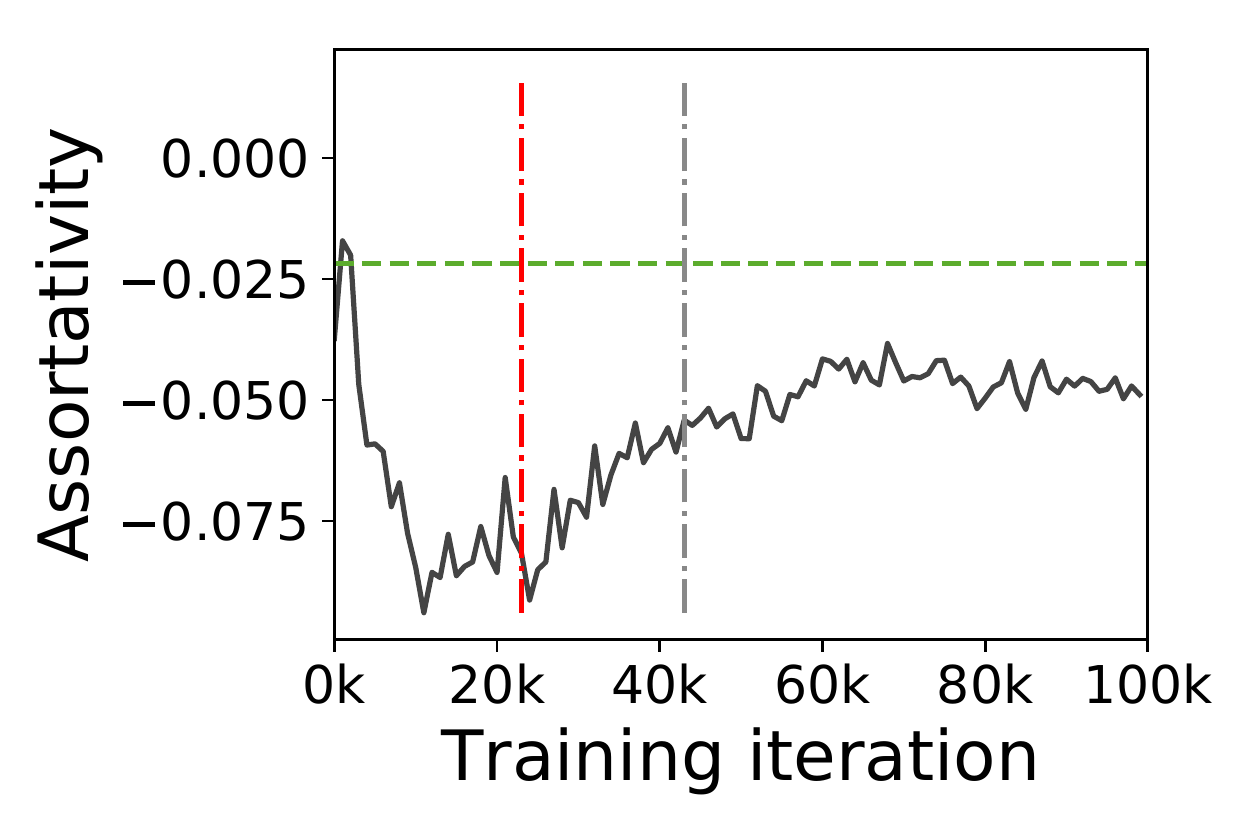}
	\caption{}
	\end{subfigure}
	\begin{subfigure}{0.32 \textwidth}
	\includegraphics[width=\textwidth]{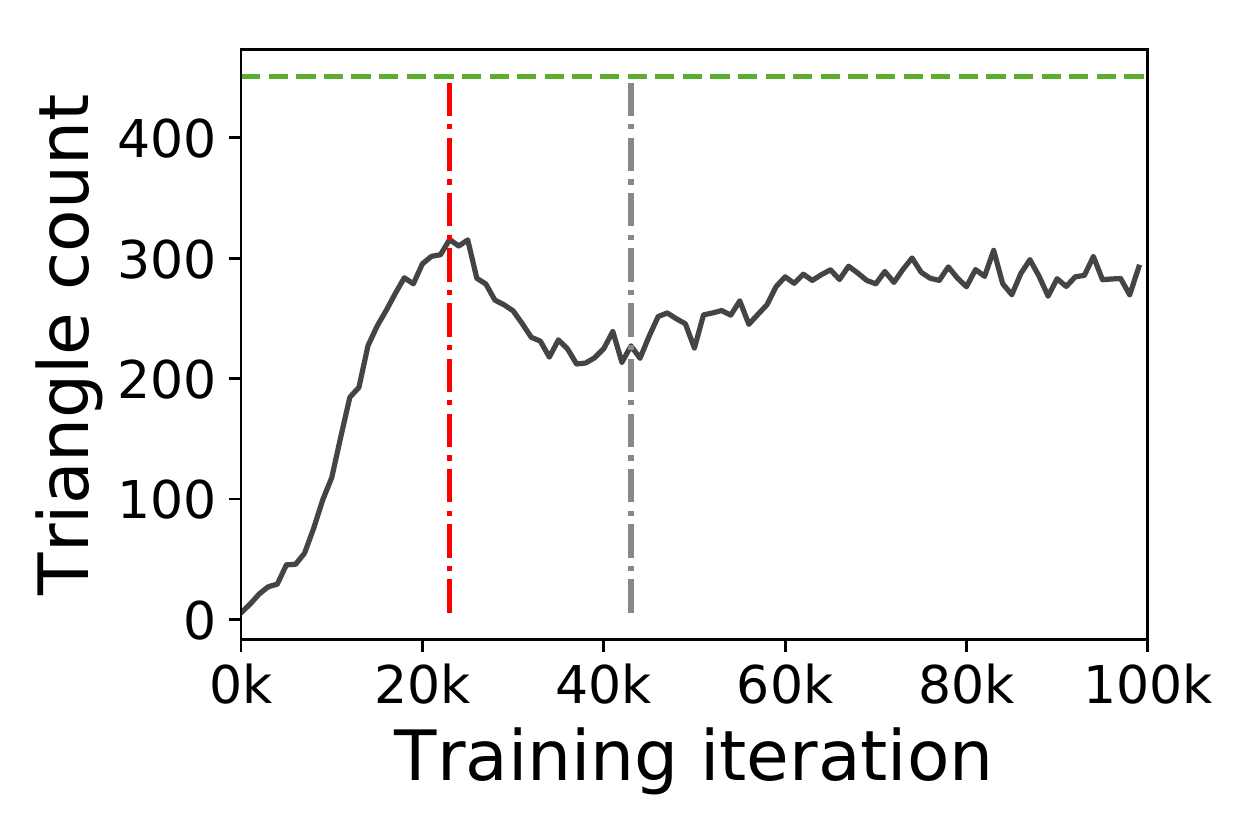}
	\caption{}
	\end{subfigure}
	
	\begin{subfigure}{0.32 \textwidth}
	\includegraphics[width=\textwidth]{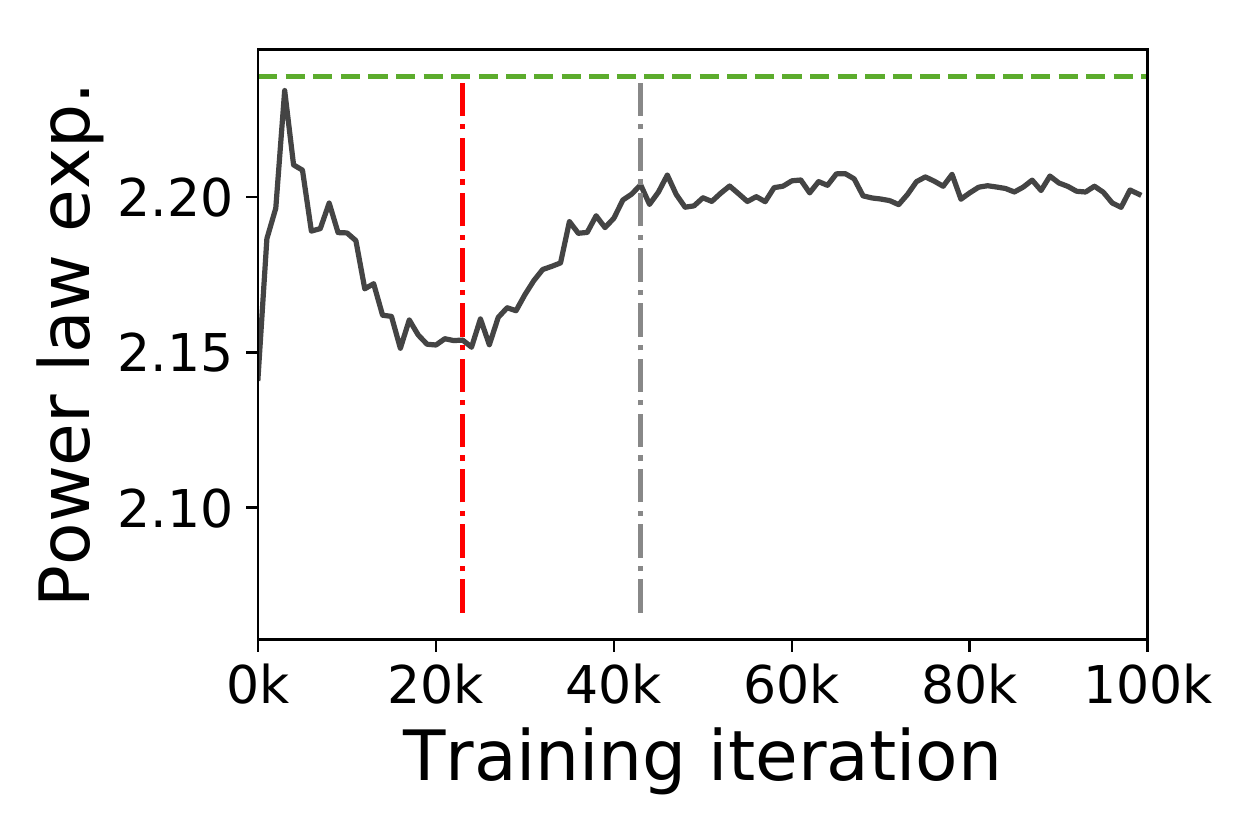}
	\caption{}
	\end{subfigure}
	\begin{subfigure}{0.32 \textwidth}
	\includegraphics[width=\textwidth]{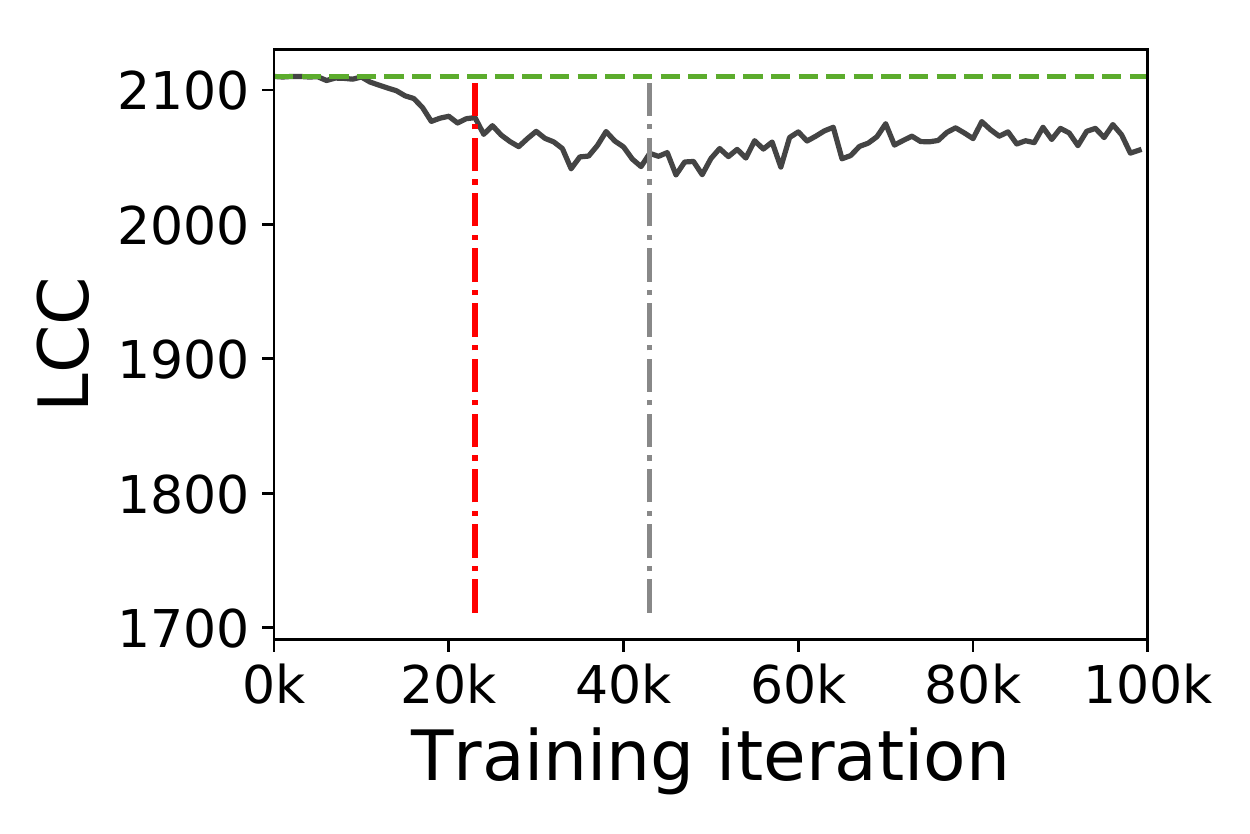}
	\caption{}
	\end{subfigure}
	\begin{subfigure}{0.32 \textwidth}
	\includegraphics[width=\textwidth]{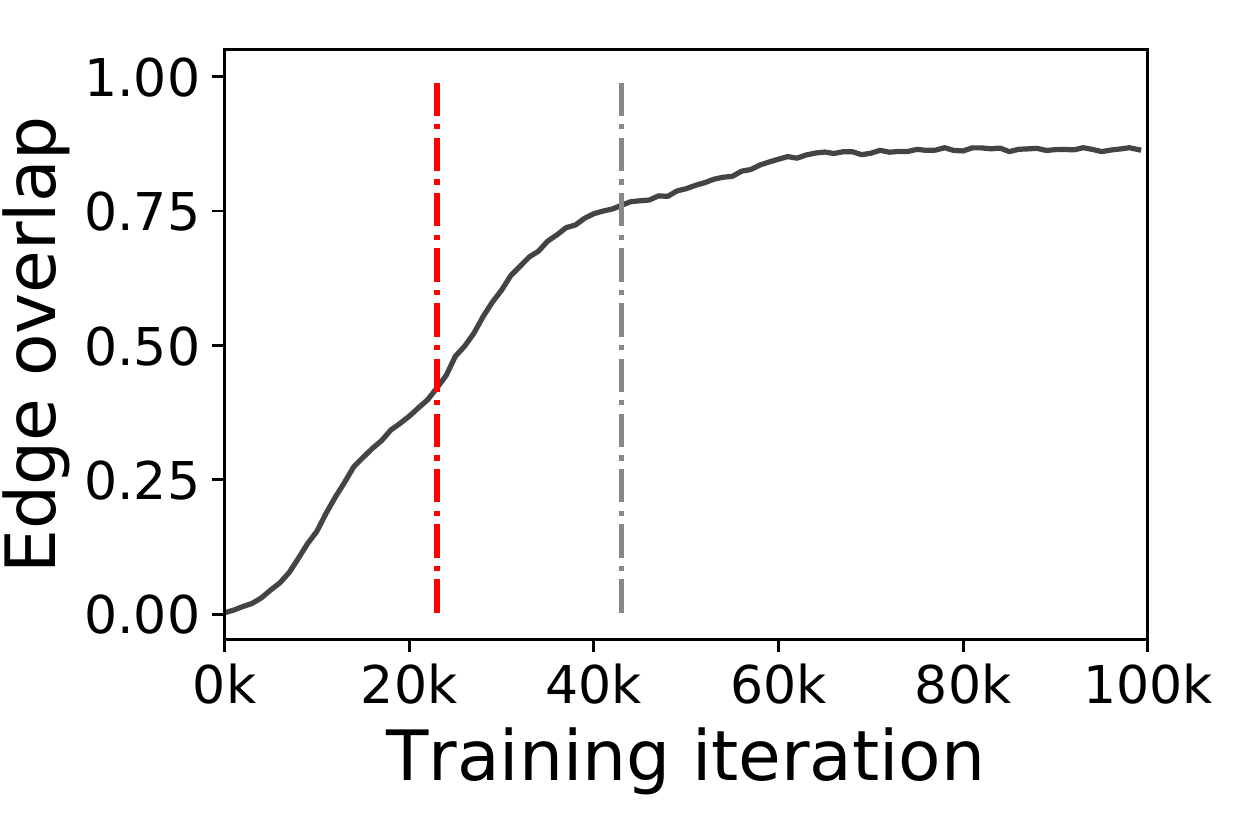}
	\caption{}
	\end{subfigure}
	\caption{Evolution of graph statistics during training on \textsc{Citeseer}}
\label{fig:app_degree_distribution_citeseer}
\end{figure*}

\begin{figure*}
\section{Latent space interpolation heatmaps}
				
\captionsetup[subfigure]{justification=centering}
\centering
	\begin{subfigure}{0.24 \textwidth}
		\includegraphics[height=0.13 \textheight]{figures/z_exploration_cora/starting_degrees}	
		\caption{Avg. degree\\of start node}
	\end{subfigure}
	\begin{subfigure}{0.24 \textwidth}
		\includegraphics[height=0.13 \textheight]{figures/z_exploration_cora/same_as_first_community}	
		\caption{Avg. share of nodes in the same comm. as the starting node}
	\end{subfigure}
	\begin{subfigure}{0.24 \textwidth}
		\includegraphics[height=0.13 \textheight]{figures/z_exploration_cora/gini}
		\caption{Gini coefficient \\(input graph: 0.48)}	
	\end{subfigure}
	\begin{subfigure}{0.24 \textwidth}
		\includegraphics[height=0.13 \textheight]{figures/z_exploration_cora/d_max}	
		\caption{Max. degree \\(input graph: 240)}
	\end{subfigure}
	
	\begin{subfigure}{0.24 \textwidth}
		\includegraphics[height=0.13 \textheight]{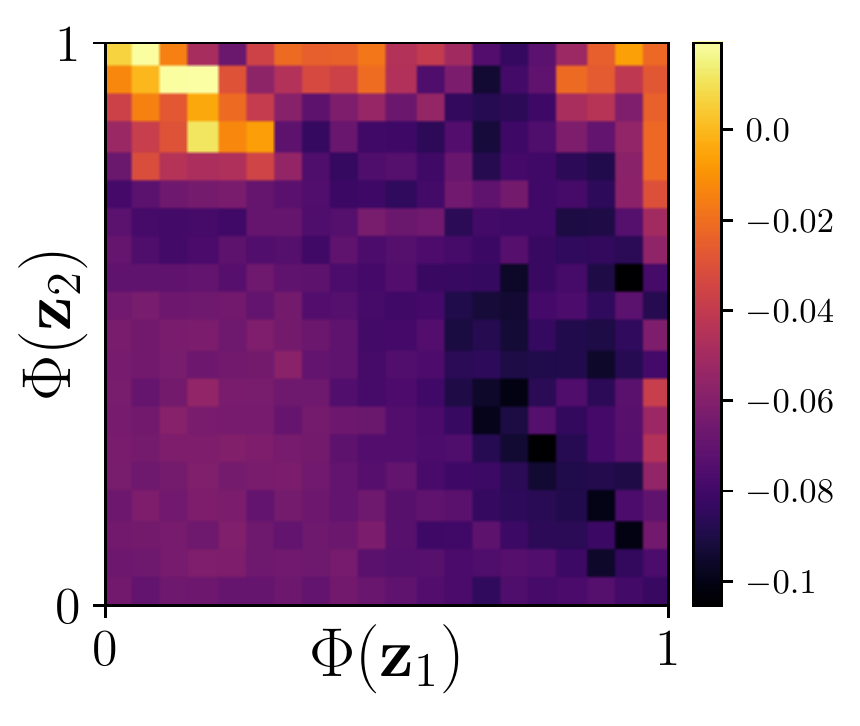}	
		\caption{Assortativity \\ (input graph: -0.075)}
	\end{subfigure}
	\begin{subfigure}{0.24 \textwidth}
		\includegraphics[height=0.13 \textheight]{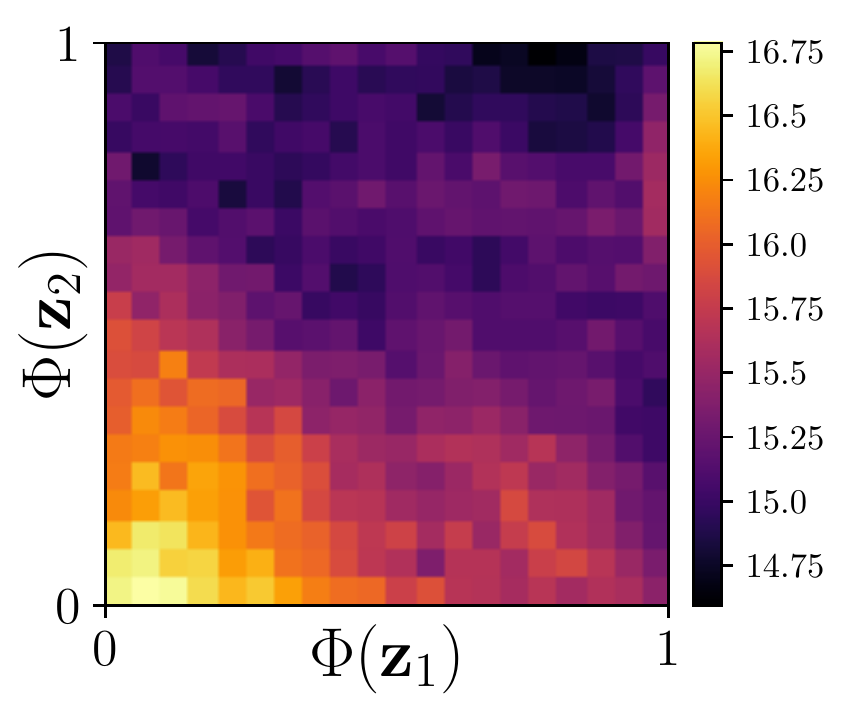}	
		\caption{Claw count \\(input graph: $3.1\times 10^{6}$)}
	\end{subfigure}
	\begin{subfigure}{0.24 \textwidth}
		\includegraphics[height=0.13 \textheight]{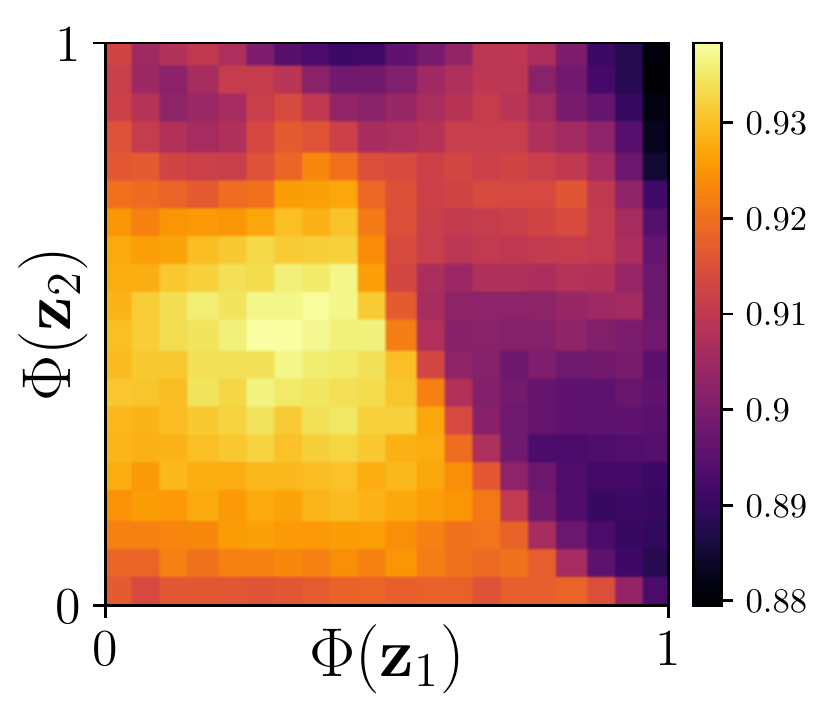}
		\caption{Rel. edge distr. entro-\\py (input graph: 0.94)}	
	\end{subfigure}
	\begin{subfigure}{0.24 \textwidth}
		\includegraphics[height=0.13 \textheight]{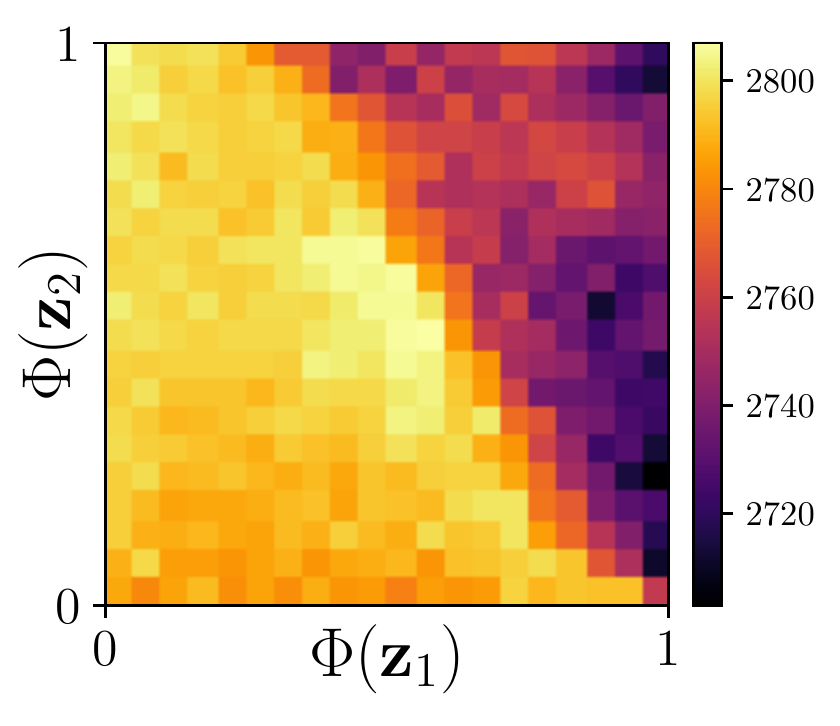}	
		\caption{Largest conn. comp. \\(input graph: 2,810)}
	\end{subfigure}
	
	\begin{subfigure}{0.24 \textwidth}
		\includegraphics[height=0.13 \textheight]{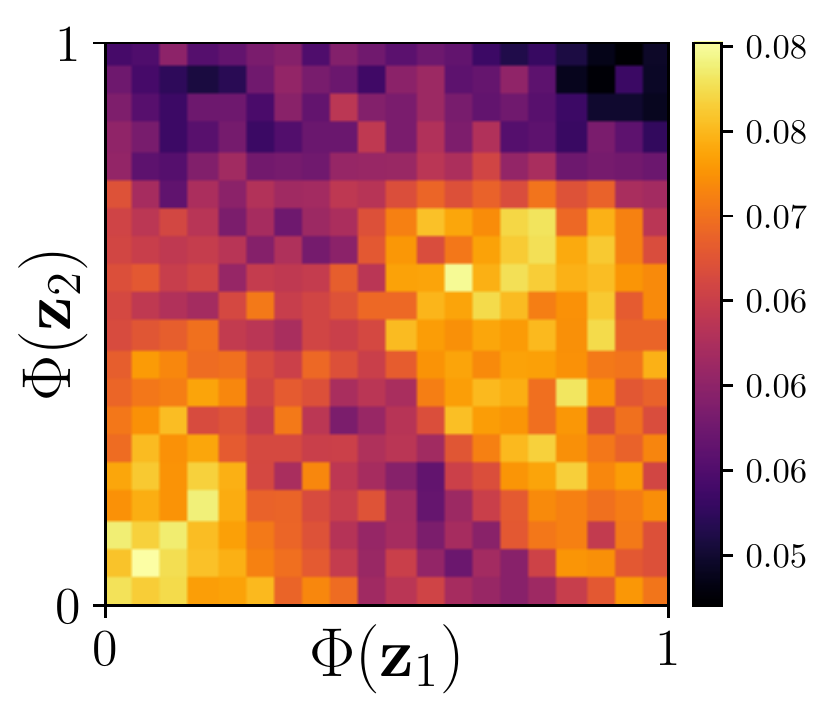}	
		\caption{Edge\\overlap}
	\end{subfigure}
	\begin{subfigure}{0.24 \textwidth}
		\includegraphics[height=0.13 \textheight]{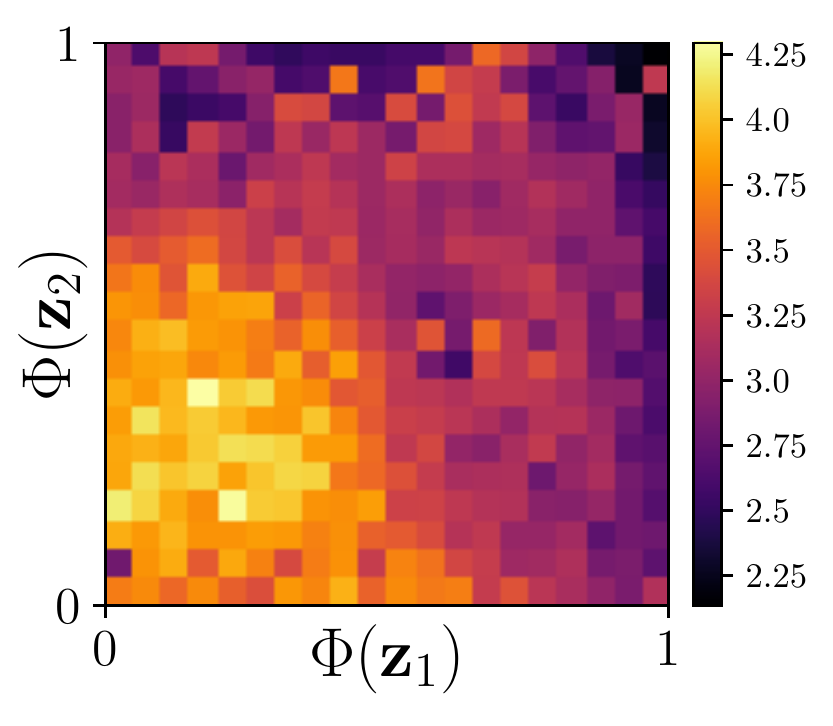}	
		\caption{Power law exponent \\(input graph: 1.86)}
	\end{subfigure}
	\begin{subfigure}{0.24 \textwidth}
		\includegraphics[height=0.13 \textheight]{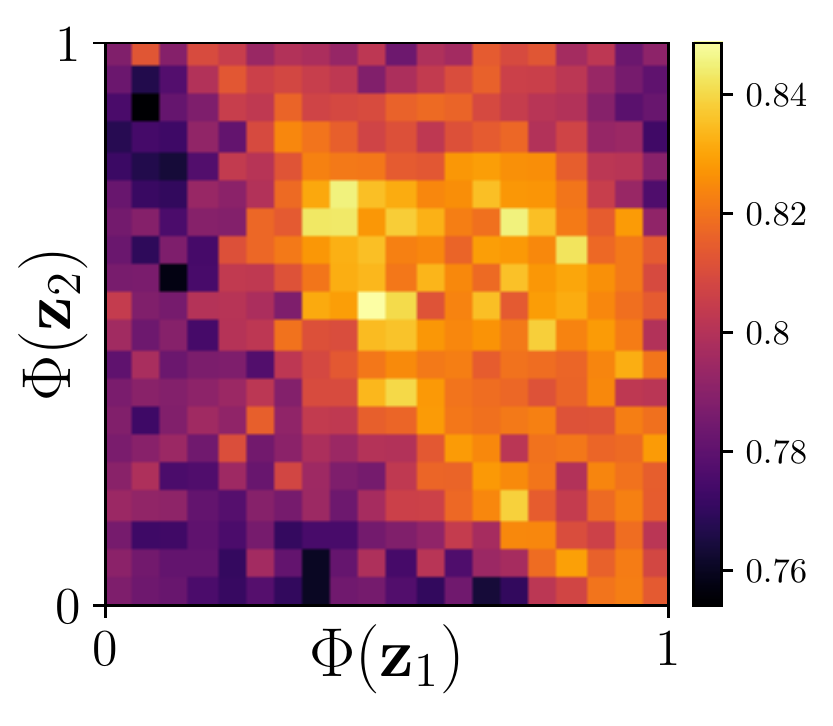}
		\caption{Avg. precision \\link prediction}	
	\end{subfigure}
	\begin{subfigure}{0.24 \textwidth}
		\includegraphics[height=0.13 \textheight]{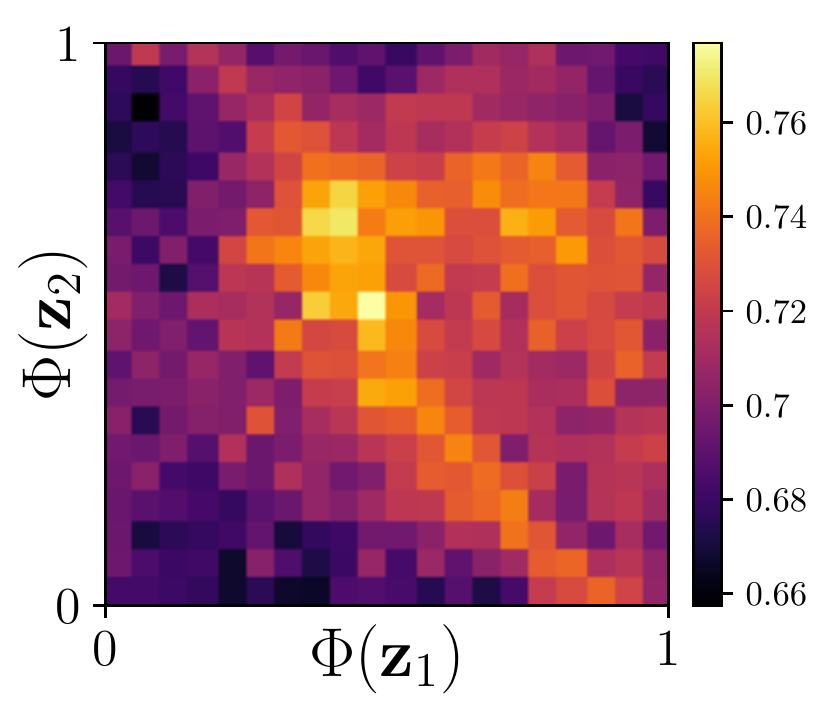}	
		\caption{ROC AUC \\link prediction}
	\end{subfigure}
	
		\begin{subfigure}{0.24 \textwidth}
		\includegraphics[height=0.13 \textheight]{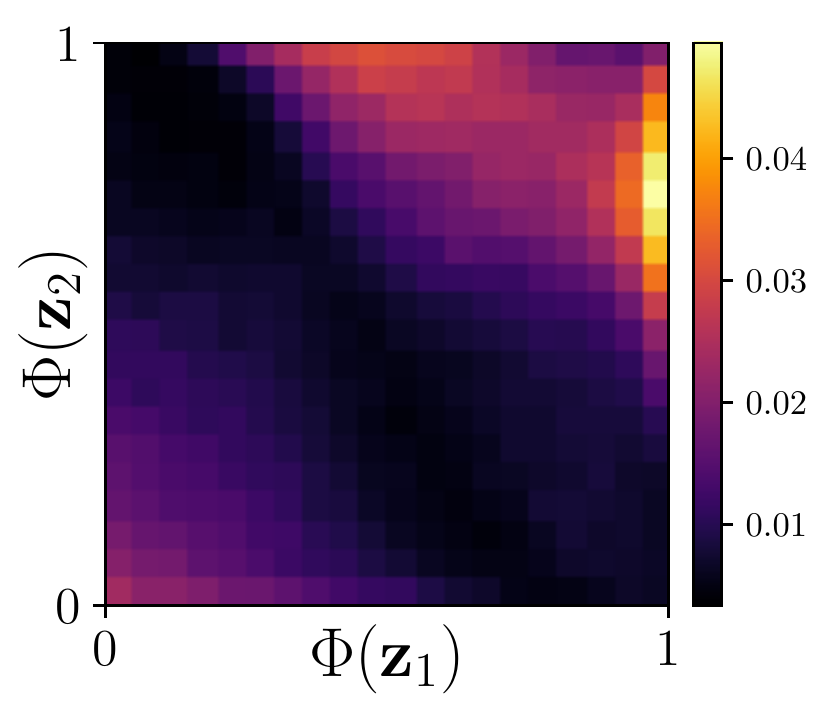}	
		\caption{Share of walks \\in single community}
	\end{subfigure}
	\begin{subfigure}{0.24 \textwidth}
		\includegraphics[height=0.13 \textheight]{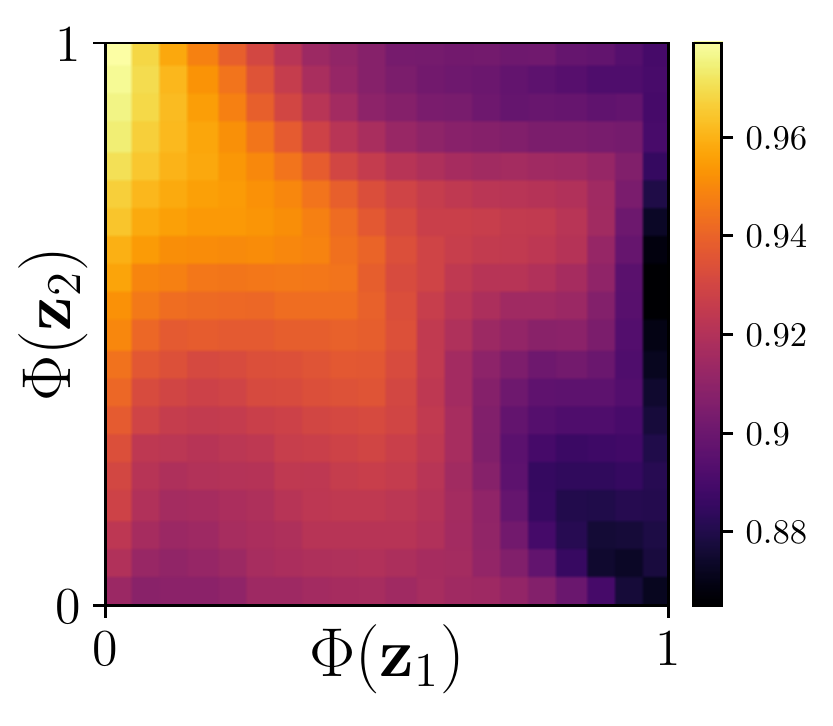}	
		\caption{tAvg. start node entropy}
	\end{subfigure}
	\begin{subfigure}{0.24 \textwidth}
		\includegraphics[height=0.13 \textheight]{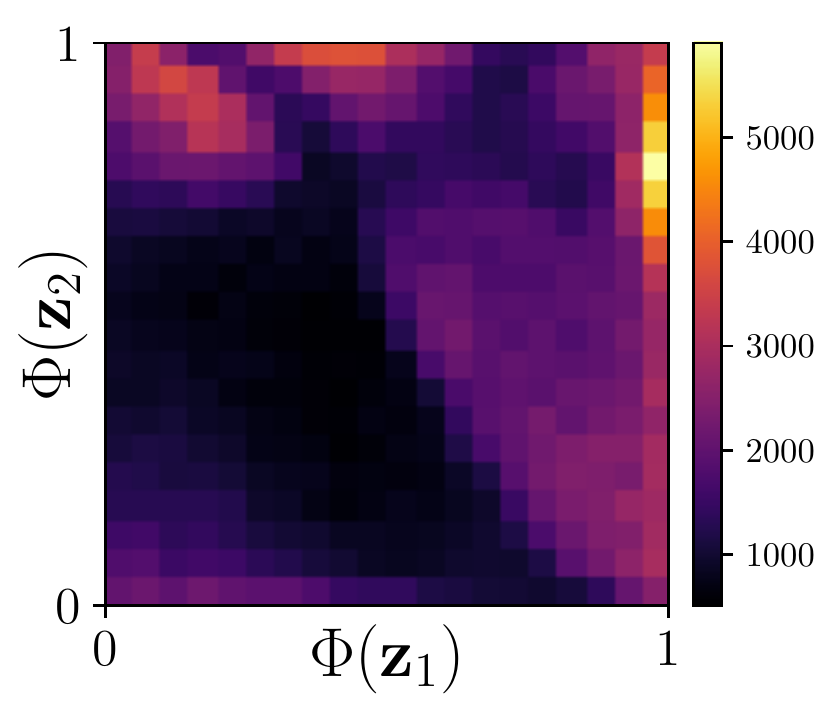}
		\caption{Triangle count \\(input graph: 2,814)}	
	\end{subfigure}
	\begin{subfigure}{0.24 \textwidth}
		\includegraphics[height=0.13 \textheight]{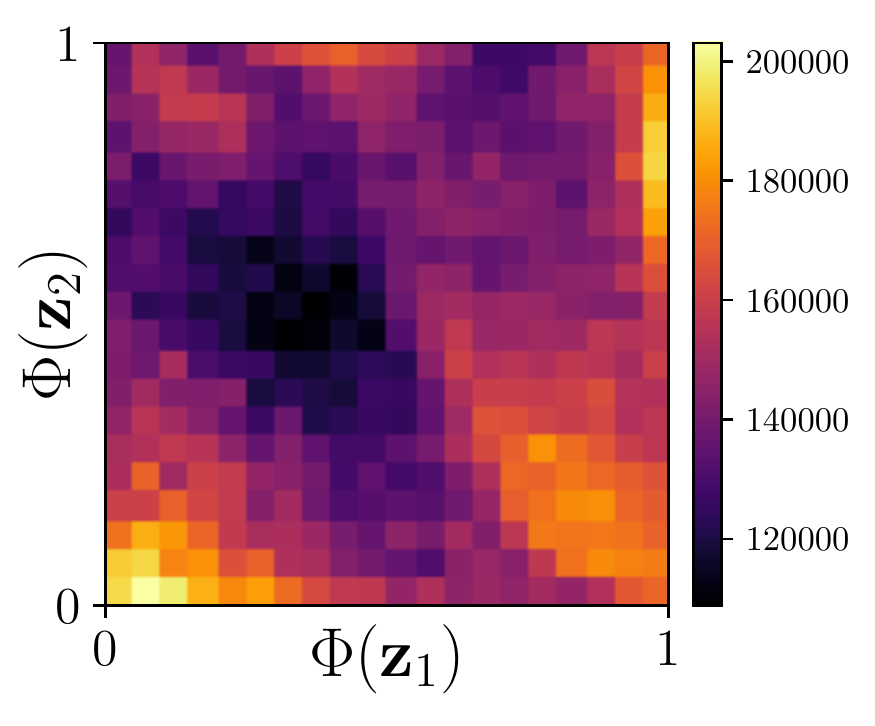}	
		\caption{Wedge count \\(input graph: 101,872)}
	\end{subfigure}

\caption{Properties of the random walks as well as the graphs sampled from the $20 \times 20$ latent space bins, trained on \textsc{Cora-ML}.}
\label{fig:app_graph_props_cora}
\end{figure*}

\begin{figure*}
	
\captionsetup[subfigure]{justification=centering}
\centering
	\begin{subfigure}{0.24 \textwidth}
		\includegraphics[height=0.13 \textheight]{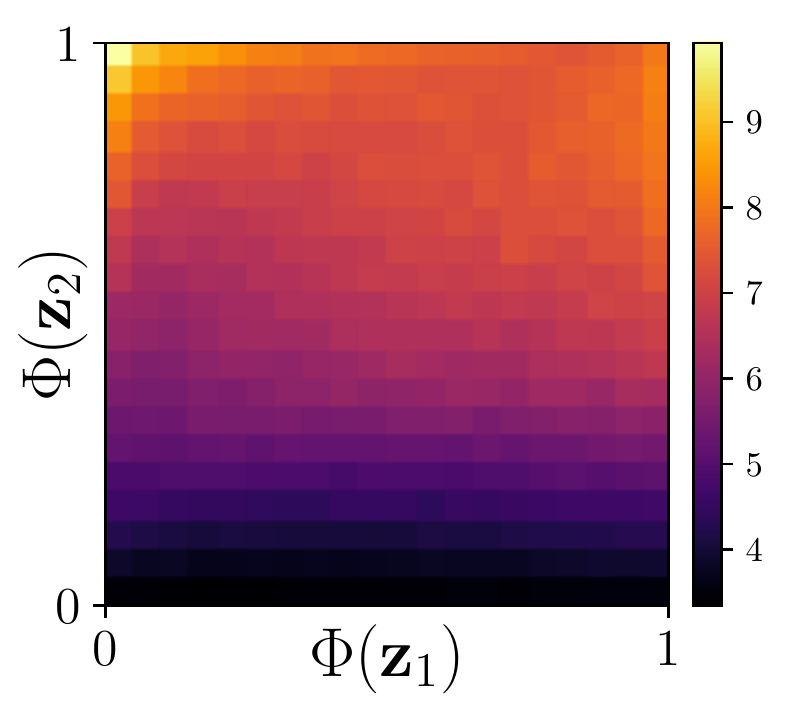}	
		\caption{Avg. degree\\of start node}
	\end{subfigure}
	\begin{subfigure}{0.24 \textwidth}
		\includegraphics[height=0.13 \textheight]{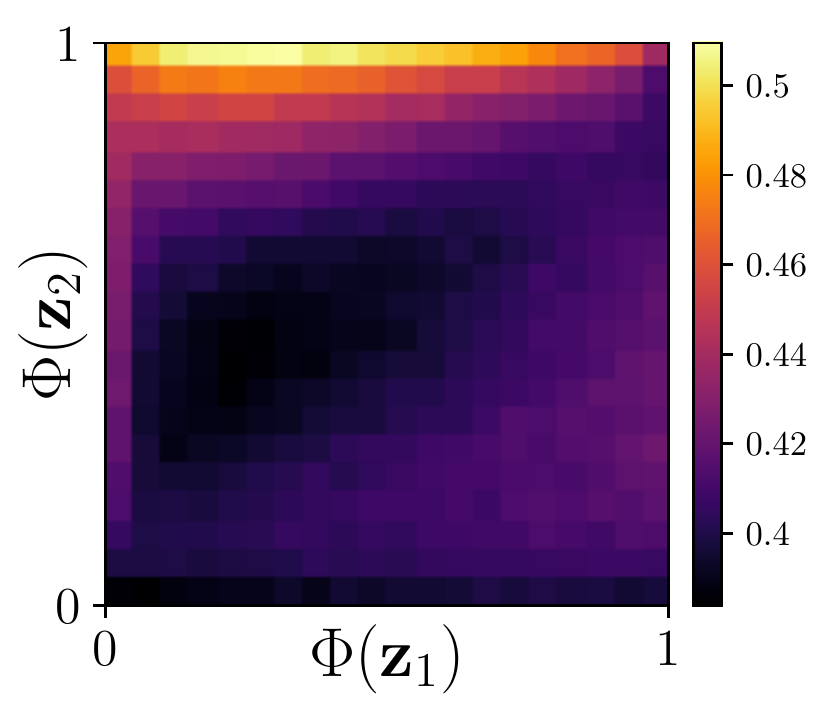}	
		\caption{Avg. share of nodes in start community}
	\end{subfigure}
	\begin{subfigure}{0.24 \textwidth}
		\includegraphics[height=0.13 \textheight]{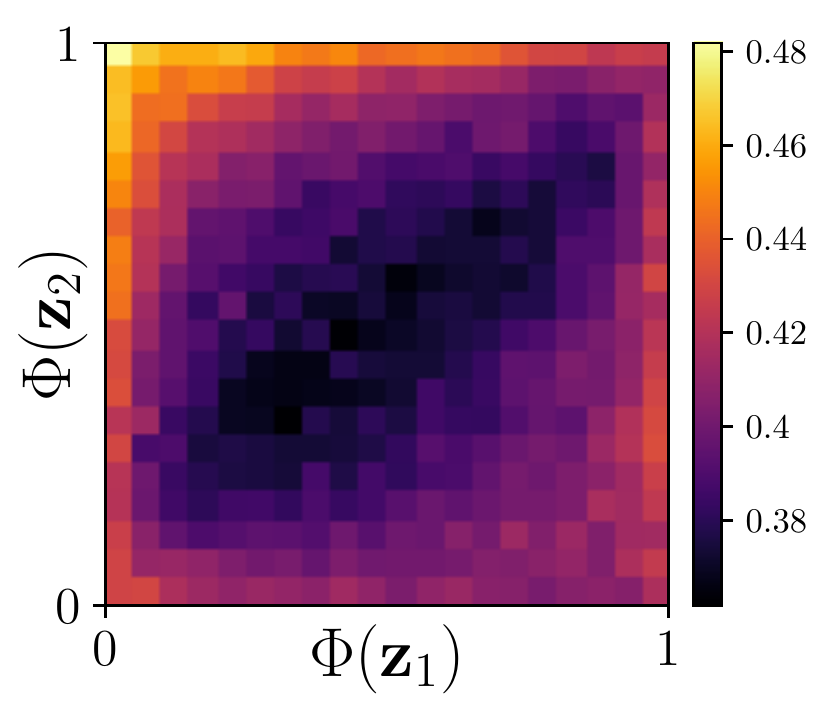}
		\caption{Gini coefficient \\(input graph: 0.404)}	
	\end{subfigure}
	\begin{subfigure}{0.24 \textwidth}
		\includegraphics[height=0.13 \textheight]{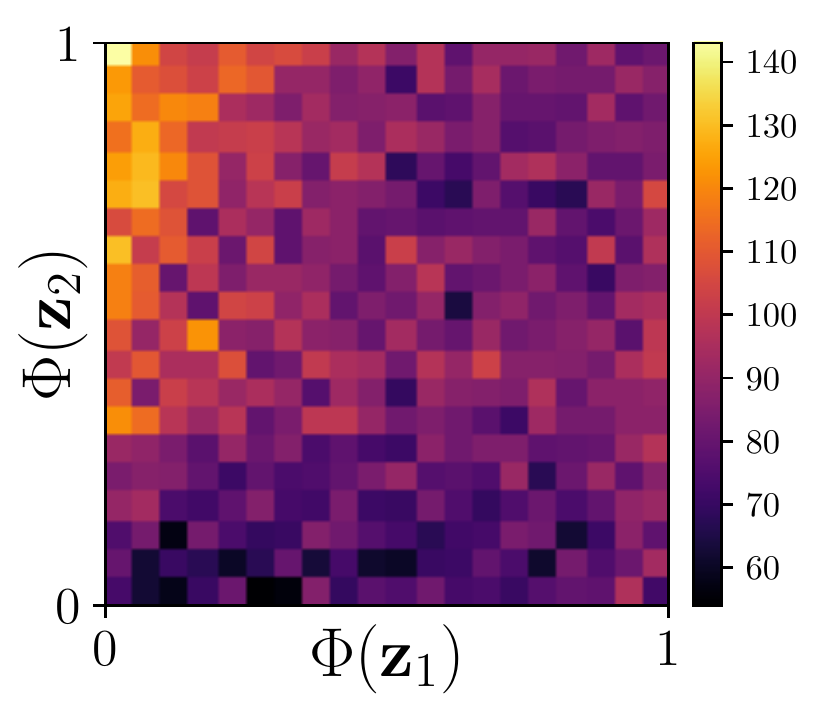}	
		\caption{Max. degree \\(input graph: 77)}
	\end{subfigure}
	
	\begin{subfigure}{0.24 \textwidth}
		\includegraphics[height=0.13 \textheight]{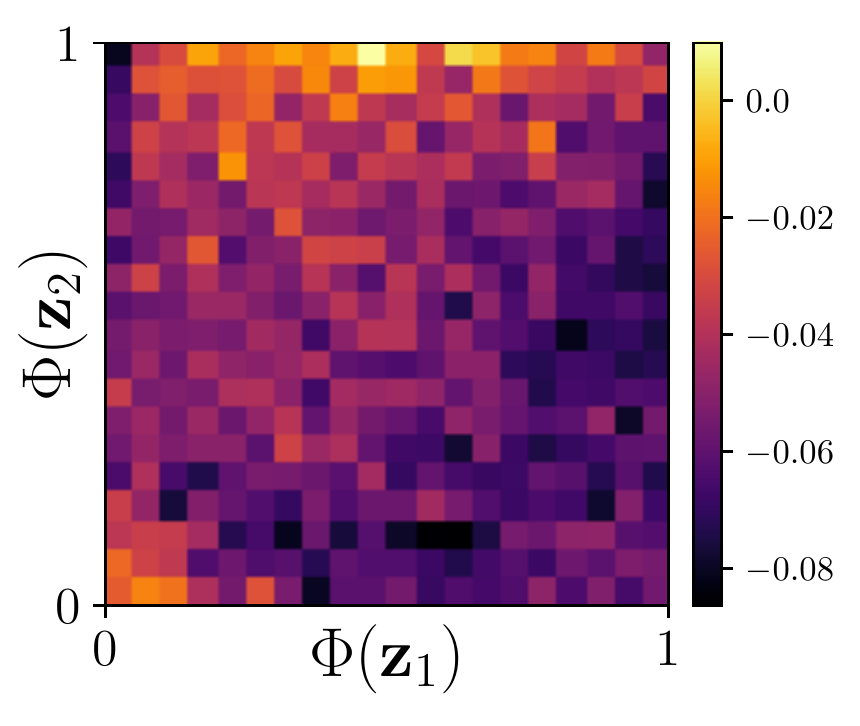}	
		\caption{Assortativity \\ (input graph: -0.022)}
	\end{subfigure}
	\begin{subfigure}{0.24 \textwidth}
		\includegraphics[height=0.13 \textheight]{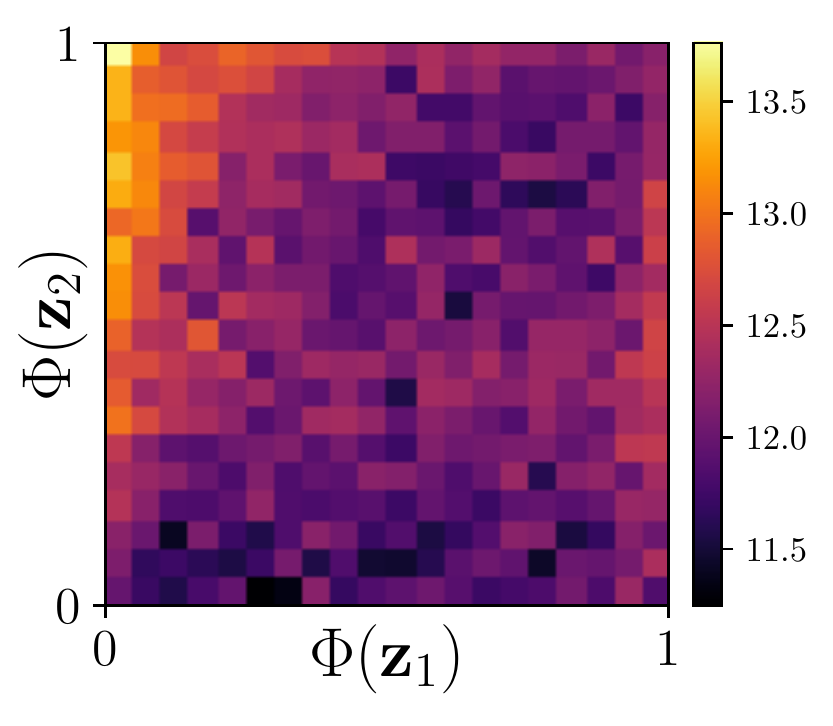}	
		\caption{Claw count \\(input graph: 125,701)}
	\end{subfigure}
	\begin{subfigure}{0.24 \textwidth}
		\includegraphics[height=0.13 \textheight]{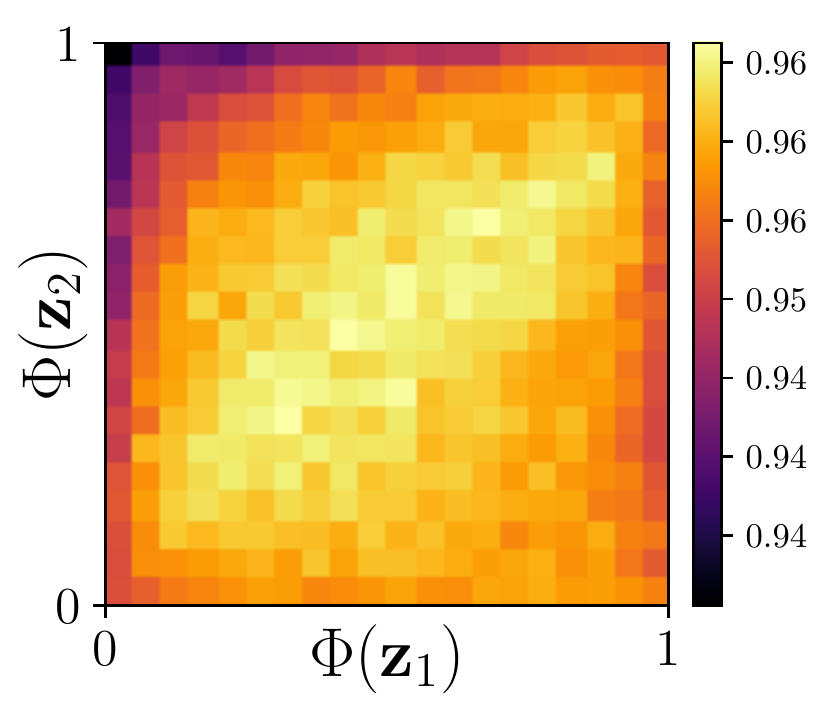}
		\caption{Rel. edge distr. entro-\\py (input graph: 0.96)}	
	\end{subfigure}
	\begin{subfigure}{0.24 \textwidth}
		\includegraphics[height=0.13 \textheight]{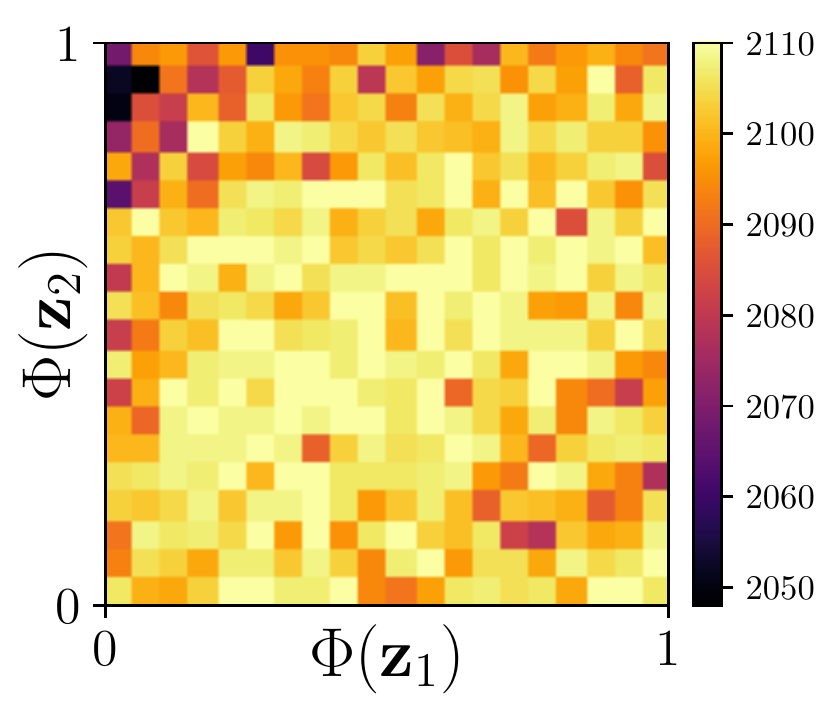}	
		\caption{Largest conn. comp. \\(input graph: 2,110)}
	\end{subfigure}
	
	\begin{subfigure}{0.24 \textwidth}
		\includegraphics[height=0.13 \textheight]{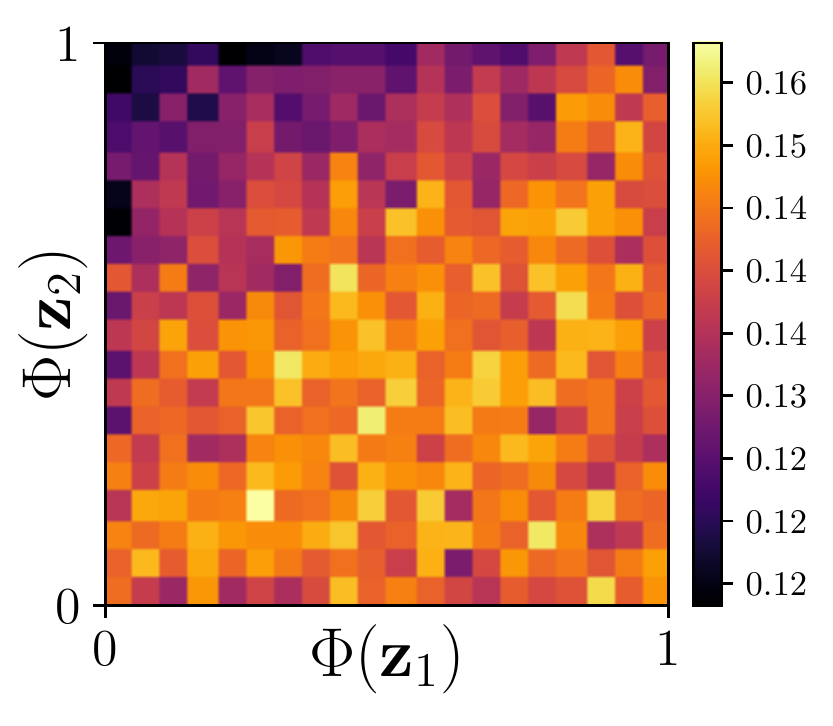}	
		\caption{Edge\\overlap}
	\end{subfigure}
	\begin{subfigure}{0.24 \textwidth}
		\includegraphics[height=0.13 \textheight]{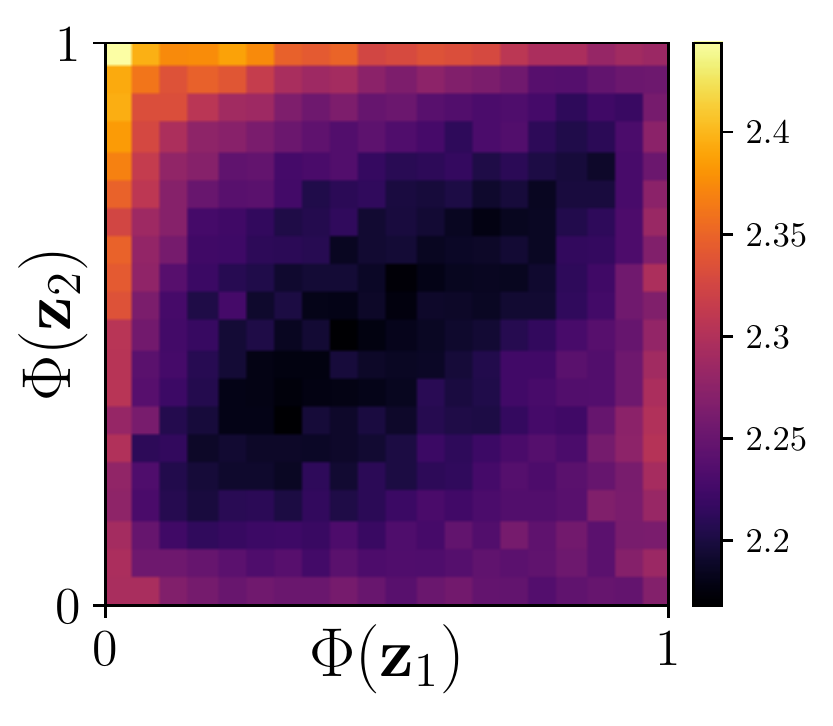}	
		\caption{Power law exponent \\(input graph: 2.239)}
	\end{subfigure}
	\begin{subfigure}{0.24 \textwidth}
		\includegraphics[height=0.13 \textheight]{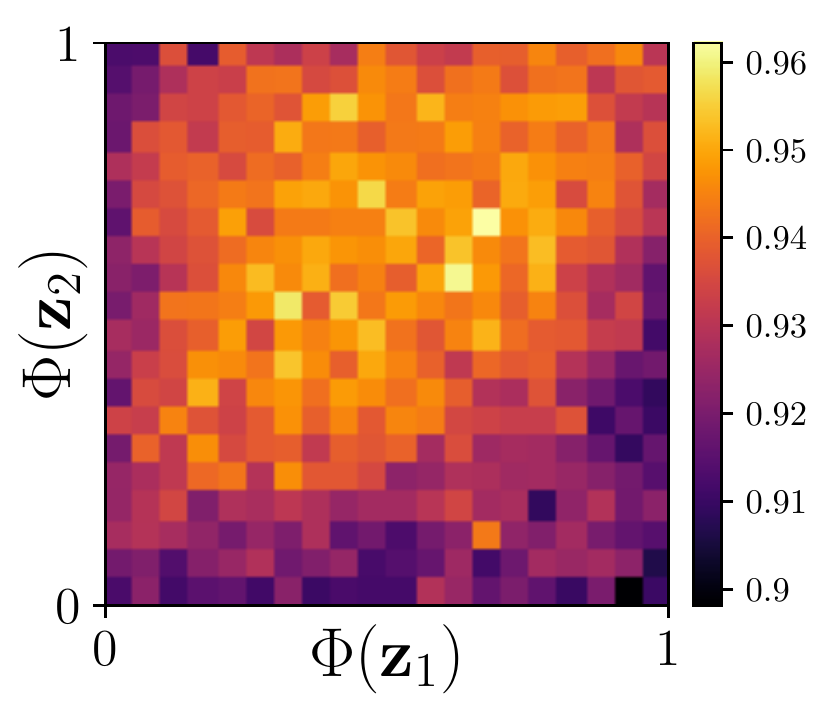}
		\caption{Avg. precision \\link prediction}	
	\end{subfigure}
	\begin{subfigure}{0.24 \textwidth}
		\includegraphics[height=0.13 \textheight]{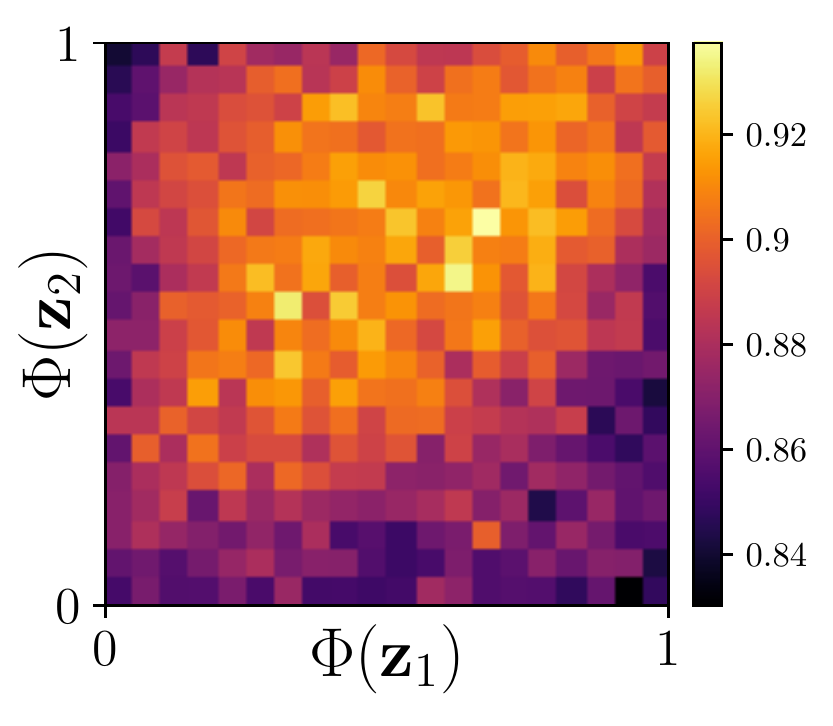}	
		\caption{ROC AUC \\link prediction}
	\end{subfigure}
	
		\begin{subfigure}{0.24 \textwidth}
		\includegraphics[height=0.13 \textheight]{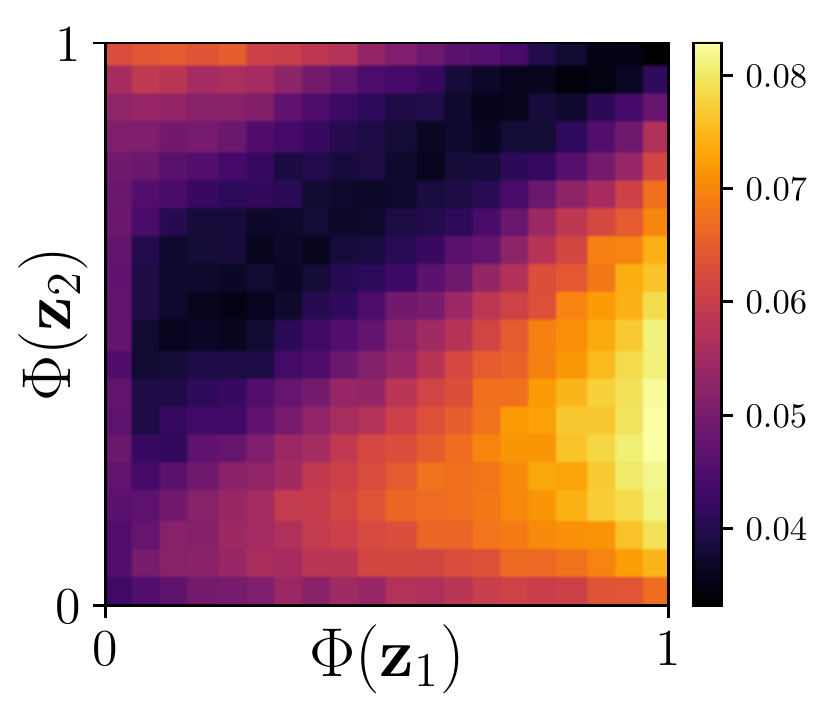}	
		\caption{Share of walks \\in single community}
	\end{subfigure}
	\begin{subfigure}{0.24 \textwidth}
		\includegraphics[height=0.13 \textheight]{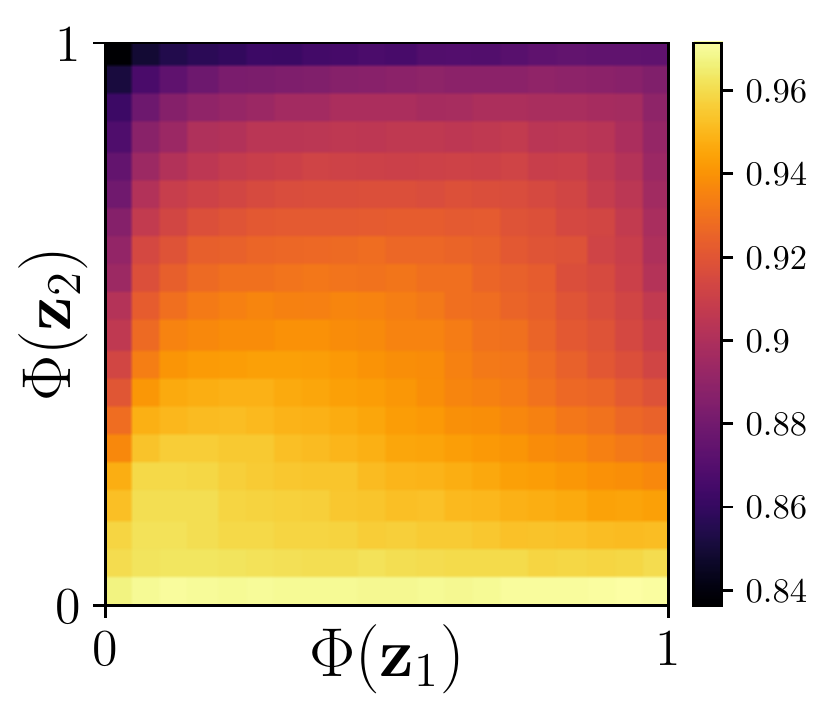}	
		\caption{Avg. start node entropy}
	\end{subfigure}
	\begin{subfigure}{0.24 \textwidth}
		\includegraphics[height=0.13 \textheight]{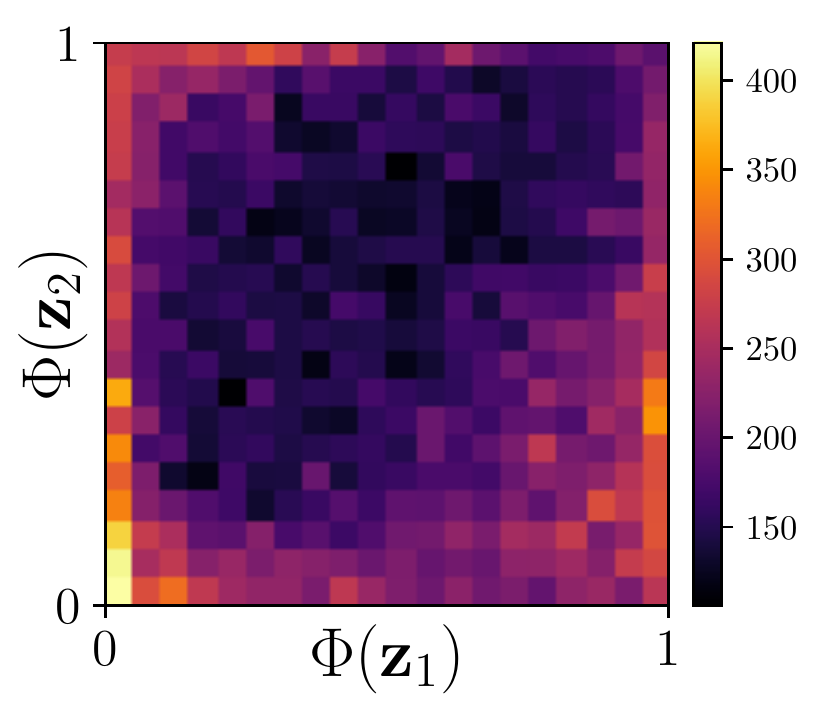}
		\caption{Triangle count \\(input graph: 451)}	
	\end{subfigure}
	\begin{subfigure}{0.24 \textwidth}
		\includegraphics[height=0.13 \textheight]{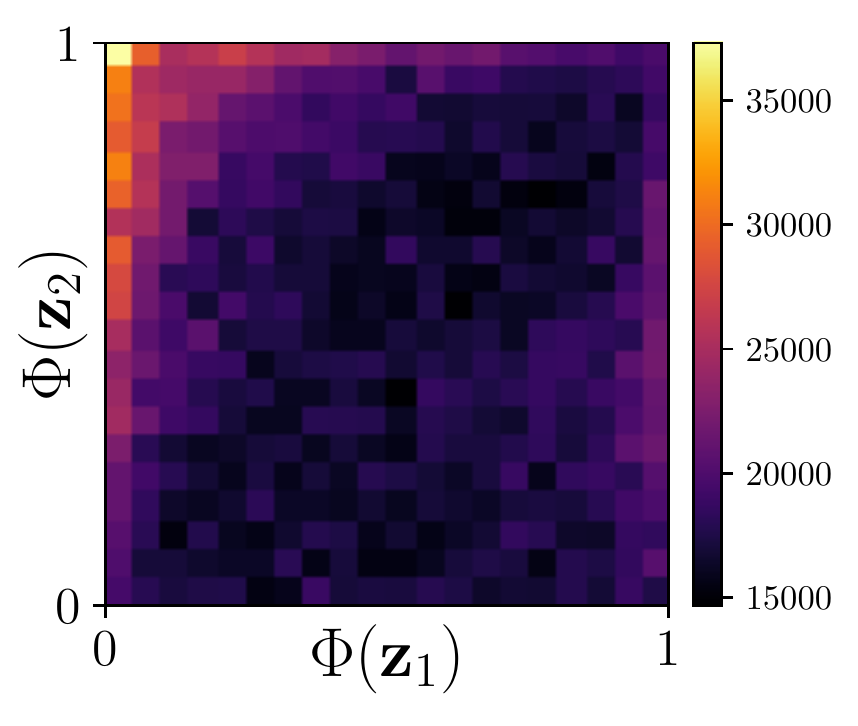}	
		\caption{Wedge count \\(input graph: 16,824)}
	\end{subfigure}

\caption{Properties of the random walks as well as the graphs sampled from the $20 \times 20$ latent space bins, trained on \textsc{Citeseer}.}
\label{fig:app_graph_props_citeseer}
\end{figure*}

\begin{figure*}[h]
\section{Latent space interpolation community histrograms -- \textsc{Citeseer}}
					\centering
	\includegraphics[width=0.6 \textwidth]{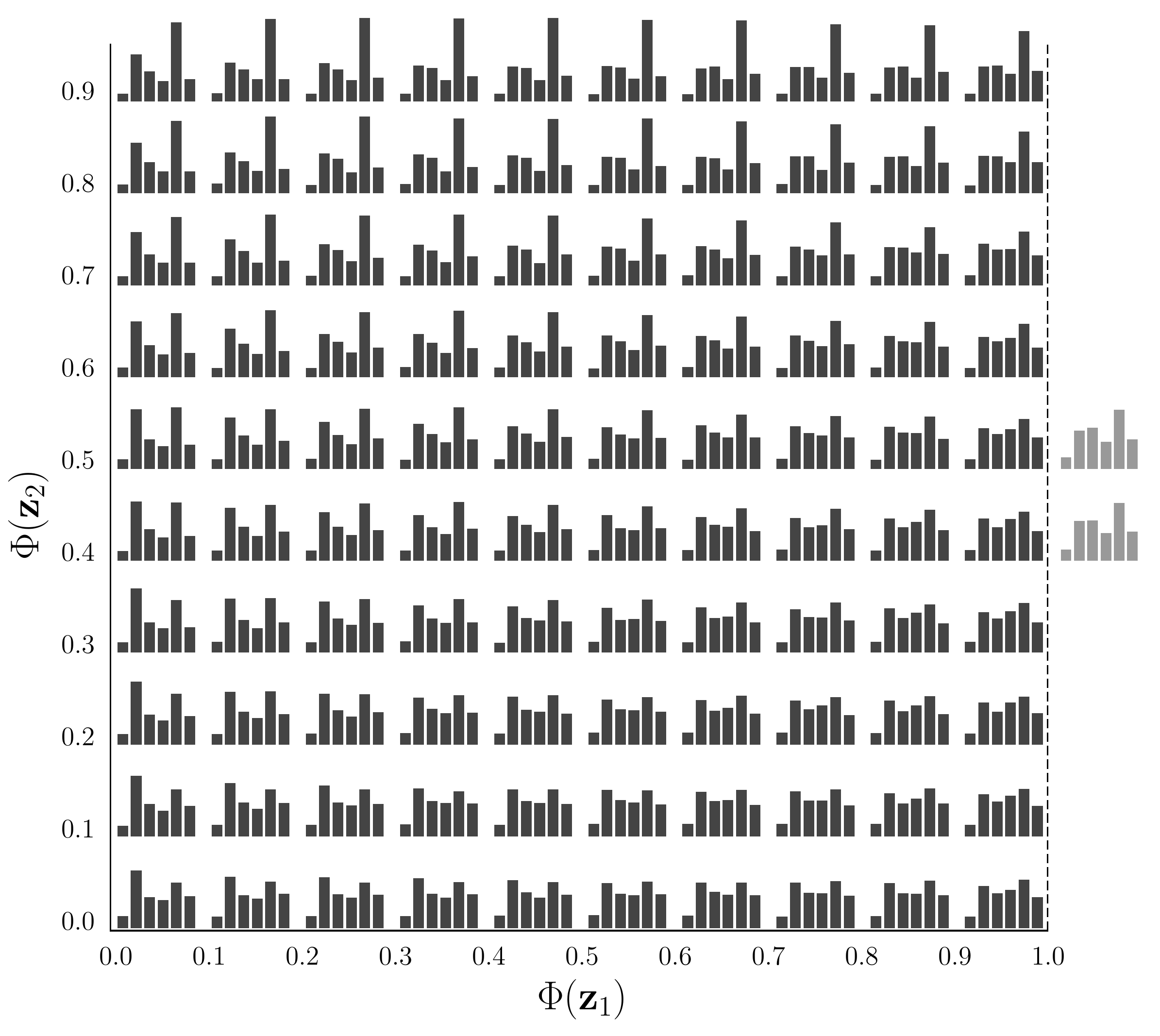}
	\caption{Community distributions of graphs generated by \name on subregions of the latent space $\bs z$, trained on the \textsc{Citeseer} network.}
	\label{fig:app_community_distr_citeseer}
\end{figure*}
\begin{figure*}
	\section{Recovering ground-truth edge probabilities}
	To further investigate the ability of \name to capture the graph structure we perform an additional experiment with the goal of analyzing how well we can recover the ground-truth edge probabilities given a graph generated from a prescribed generative model.
	Towards that end, first, we generate a graph from \SBM ($N=300$ nodes and $3$ communities), then we fit NetGAN on this graph, and finally we compare the ground truth edge probabilities to the edge scores inferred by NetGAN -- specifically we compute their ranking correlation. We find a correlation of 0.998 (with EO = 0.42), which shows that NetGAN uncovered the underlying generative process, without overfitting to the input graph.
\end{figure*}
\begin{figure*}
	\section{Hyperparameter configuration}
	As discussed in Sec.\ \ref{sec:link_prediction} \name is not sensitive to the choice of most hyperparameters. For completeness, we report here sensible defaults that we used in used in our experiments.
	The generator and discriminator each have a single hidden layer with $40$ and $30$ hidden units respectively. The down-projection matrix for the generator is $\bs W_{down, g} \in \mathbb{R}^{N \times H_g}$ with $H_g=64$, and for the discriminator is $\bs W_{down, d} \in \mathbb{R}^{N \times H_d}$ with $H_d=32$.
	The latent code $\bs z$ is drawn from a $d=16$ dimensional multivariate standard normal distribution.
	We anneal the temperature from $\tau=1.0$ down to $\tau=0.5$ every 500 iterations with a multiplicative decay of $0.995$. We tune the parameters $p$ and $q$ (used to bias the generated random walks) for each dataset separately using the procedure in \citet{grover2016node2vec}.
	
	\smallskip
	We use Adam \cite{kingma2014adam} to optimize all the parameters with a learning rate of $1e{-3}$ and we set the regularization strength for the $L_2$ penalty to $1e{-6}$. 
	We perform five update steps for the parameters of the discriminator for each single update step of the parameters of the generator, and we set the Wasserstein gradient penalty applied to the discriminator to $10$ as suggested by \citet{gulrajani2017improved}.
	For early stopping, we evaluate the score every 500 iterations, and set the patience to 5 evaluation steps. To calculate the validation score we generate 15M transitions, e.g. for a random walk of length 16 (i.e. 15 transitions per random walk) this equals 1M random walks.

	\smallskip
	For more details we refer the reader to the provided reference implementation at \href{https://www.kdd.in.tum.de/netgan}
	{https://www.kdd.in.tum.de/netgan}.
\end{figure*}

\end{document}